Statistical NLP for Optimization of Clinical Trial Success Prediction in Pharmaceutical R&D

by Michael Robert Doane

B.S. in Chemical Engineering, May 2019, University of Massachusetts – Lowell
M.S. in Bioengineering, December 2021, Stanford University

A Praxis submitted to

The Faculty of
The School of Engineering and Applied Science
of The George Washington University
in partial fulfillment of the requirements
for the degree of Doctor of Engineering

August 15, 2025

Praxis directed by

Everett Oliver
Professorial Lecturer of Engineering Management and Systems Engineering

The School of Engineering and Applied Science of The George Washington University certifies that Michael Robert Doane has passed the Final Examination for the degree of Doctor of Engineering as of July 23, 2025. This is the final and approved form of the Praxis.

**Statistical NLP for Optimization of Clinical Trial Success Prediction in Pharmaceutical R&D**
Michael R. Doane

Praxis Research Committee:

  Everett Oliver, Professorial Lecturer of Engineering and Applied Science, Praxis Director

  Stanley Small, Professorial Lecturer of Engineering and Applied Science, Committee Chair

  Daniel Veit, Professorial Lecturer of Engineering and Applied Science, Committee Member

# Dedication

First and foremost, this work is dedicated to my loving wife, Amanda, whose patience and belief in me enabled me to persevere through the challenges of this program. May all who read this be fortunate enough to know someone as kind and compassionate as her.

To my mother and father, Catherine and Dennis, who instilled in me from a very early age the value of education. They recognized academia as a path that could lead to a rich and fulfilling life, and I am eternally grateful for their sacrifices and encouragement.

To my closest friends, Ian, Josh, Ag Mohd, Steven, and Chris, who have been there to celebrate with me in the good times, and shepherd me through the tough times.

Finally, I have been fortunate enough to have several great teachers and mentors in my life. Scott Hildreth and Bruce Mayer of Chabot College taught me to push myself academically beyond what I had thought I was capable of. Dr. Prakash Rai of UMass Lowell believed in me before I knew whether I should believe in myself. Finally, in business, Kundini Amin inspired in me the professional confidence to pursue a shift to my current career path, which brought me to the subject matter explored in this praxis.

Without the people whose names are mentioned above, this work would not exist.

## Acknowledgements

I would like to acknowledge the tireless efforts of my advisor, Dr. Everett Oliver, whose guidance and encouragement was critical in the development of this work. Always there to make a suggestion, be a sounding board for new ideas, or just to lend an empathetic ear to frustration, he exemplifies what academic advising should be.

This scholastic milestone would not have been possible without generous financial assistance from my employer, Biogen. I also greatly appreciate the wonderful technical inspiration I had from speaking about this work with my brilliant colleagues, Dr. Joseph Charest and Cedric Liu.

The genesis of this work was not straightforward, and along that circuitous path I encountered a great many individuals who helped to shape my appreciation and intuition around technical inquiry, probability estimation, and machine learning. Though there may be too many of you to name, please know that I do cherish your immeasurable contribution to my educational and professional journey.

# Abstract of Praxis

## Statistical NLP for Optimization of Clinical Trial Success Prediction in Pharmaceutical R&D

Pharmaceutical research and development is plagued by high attrition rates and enormous clinical trial costs, particularly within neuroscience, where overall industry success rates are below 10%. Timely identification of promising programs can streamline resource allocation and reduce financial risk. This praxis presents the development and evaluation of an NLP-enabled probabilistic classifier designed to estimate the probability of technical and regulatory success (pTRS) for clinical trials in the field of neuroscience. Leveraging data from the ClinicalTrials.gov database and success labels from the recently developed Clinical Trial Outcome dataset, the classifier extracts text-based clinical trial features using statistical NLP techniques. These features were integrated into several non-LLM frameworks (logistic regression, gradient boosting, and random forest) to generate calibrated probability scores. Model performance was assessed on a retrospective dataset of 101,145 completed clinical trials spanning 1976–2024, achieving an overall ROC-AUC of 0.64. An LLM-based predictive model was then built using BioBERT, a domain-specific language representation encoder. The BioBERT-based model achieved an overall ROC-AUC of 0.74 and a Brier Score of 0.185, indicating its predictions had, on average, 40% less squared error than would be observed using industry benchmarks. The BioBERT-based model also made trial outcome predictions that were superior than benchmark values 70% of the time, overall. By integrating NLP-driven insights into drug development decision-making, this work aims to enhance strategic planning and optimize investment allocation in neuroscience programs.

# Table of Contents

# List of Figures

# List of Tables

# List of Acronyms

| | |
|---|---|
| BERT | Bidirectional Encoder Representations from Transformers |
| CTO | Clinical Trial Outcome |
| PR-AUC | Precision Recall Area Under Curve |
| pRS | Probability of Regulatory Success |
| pTRS | Probability of Technical and Regulatory Success |
| pTS | Probability of Technical Success |
| ROC-AUC | Receiver Operating Characteristic Area Under Curve |
| R&D | Research and Development |

# Chapter 1—Introduction

## 1.1 Background

Pharmaceutical research and development represents a potentially lucrative, but very costly endeavor. When accounting for cost of capital and failed therapeutic candidates, the cost of developing a drug had risen at an annual rate 8.5% above inflation from 2003 to 2013, to $3.88 billion per new approved compound in 2024 dollars (DiMasi, et al., 2016). These cost constraints make drug failures particularly difficult for pharmaceutical firms to endure, while preventing those resources from being used for development of other potential therapeutic candidates.

This work will concern itself primarily with clinical development phases (Phase 1, 2, and 3). Drug therapies require a particularly long development timeline, with risks and unknowns becoming resolved slowly over time as studies read out. New drugs may behave very differently in human subjects than in *in vitro* or animal testing during preclinical phases, so it is very difficult for firms to assess the probabilities of technical and regulatory success (pTRS) for emerging molecular entities.

The uncertainties around these evaluations include:

- Patient safety with respect to the active ingredient(s) (the molecules that give rise to clinical therapeutic effect) and its chosen formulation (the overall composition of the drug product, including inactive ingredients; particularly relevant in Phase 1)
- Effectiveness (efficacy) of the active ingredient(s) in the clinical indication(s) for which approval is being sought (particularly relevant in Phase 2 and 3)

- Competitors may have positive trial readouts (interim or completed trial results) for drugs treating the same clinical indication, imposing a higher bar for follow-on drugs to surpass in order to gain approval (relevant in all phases)
- Dosage uncertainty presents a risk until the appropriate dosing regimen for a drug has been established in humans (Phases 1 and 2)
- Biomarker-related risks: not having a biomarker (a measurable molecule in patients that relates to disease state or outcome) to help identify ideal patient populations and evaluate molecular function can impede trial success (all phases)

As drugs progress through preclinical and clinical stages, more is learned about their scientific and clinical properties. Thus, drug development results in the discharge of technical risks such as those surrounding safety, efficacy, dosage, etc., but resolving this uncertainty requires enormous capital investment.

A chief aim of R&D organizations is to accurately characterize their drug candidate's probability of technical and regulatory success (pTRS) so that program value proposition, and value for patients, can be appropriately assessed. This can then guide enterprise decision making (whether to progress, accelerate, or terminate a program). Several methods exist for pTRS estimation, often involving the use of industry-wide success rate benchmarks. This can be combined with some form of expert elicitation, a process in which the technical experts in a product development team provide subjective input related to a program's likelihood of success, based on a myriad of factors that are difficult to quantify. However, this process is time consuming, and subject to the biases of the experts being interviewed.

**1.2 Research Motivation**

When a program that progresses to a new phase eventually fails, this unsuccessful candidate represents a poor result for both companies and patients:

1. The opportunity cost of having invested in an asset that did not succeed rather than one that might have

2. Patients are exposed to a compound that is not efficacious, and/or does not have an appropriate safety and tolerability profile

Because of this, even small improvements in the quality of pTRS predictions can yield enormous benefits for the health of both patients and the industry.

**1.3 Problem Statement**

*The inability of pharmaceutical companies to accurately scope the technical risks of R&D product development scenarios diminishes R&D investment returns, with overall drug success rates of approximately 10%.* (Zhou, 2018).

These drug failures generate capitalized cost burdens that contribute to the overall cost of modern drugs, and are passed on to payers and patients. The situation is particularly difficult in the field of neuroscience, which is notorious for having low clinical trial success rates and a lower innovation index compared to other innovative fields like immunology and oncology (Zaragoza Domingo, et al., 2024).

A challenge for predicting future outcomes of drug trials is that there have not historically been large accurate labels of whether past trials were successful. Recently, researchers at the University of Illinois – Urbana-Champaign developed a post-hoc method of assessing with high accuracy whether clinical trials that have been completed were successful based on a number of publicly available sources of information. (Gao, 2024). The database they have built, the CTO Dataset, now can be used as a single source of truth, permitting the building of supervised learning models to generate predictions of clinical trial success for ongoing or future trials.

**1.4 Thesis Statement**

*A predictive model using Natural Language Processing is needed to improve Probability of Success (pTRS) predictions for neuroscience programs in pharmaceutical R&D.*

Current means for generating pTRS are long, cumbersome, and vulnerable to the subjectivity of project teams with a vested interest in the success of the program. The ability to make better, faster, objective predictions about the success of neuroscience programs would enable enormous improvements in R&D.

**1.5 Research Objectives**

Initial steps for this work will involve extracting data from ClinicalTrials.gov (n.d.) and the trial success estimates from the CTO Dataset built by Gao, et al. (2024). These data will be integrated, filtered, and cleaned, before implementing the BERT (Bidirectional Encoder Representations from Transformers) deep learning model from Google, which will allow for the construction of a model trained on pre-2019 neuroscience trial data that is capable of generating pTRS estimates ranging from 0 to 1. Evaluation of this model will be accomplished by testing its ability to predict the outcome of neuroscience trials from after 2019 and quantifying accuracy using log loss, ROC-AUC, Brier score, and other metrics. These accuracy scores will be compared to the scores that would be seen by using industry benchmarks alone. Separate models will be built and validated for trial Phases 1, 2, and 3, and words and phrases that have a particularly strong correlation with trial outcomes will be examined.

**1.6 Research Questions and Hypotheses**

**RQ1:** Can a non-LLM NLP model be built that can generate >5% better-than-random pTRS predictions for neuroscience indications?

**RQ2:** Does the use of an LLM-enhanced NLP model lead to a >5% improvement in pTRS prediction over non-LLM for neuroscience indications?

**RQ3:** Can a model trained solely on ClinicalTrials.gov data and outcome labels surpass the predictive performance of industry benchmarks by >5% for neuroscience indications?

**RQ4:** Does use of training data from all indications (not just neuroscience) improve neuroscience pTRS prediction >5%?

**H1:** A non-LLM NLP model can be built that can generate >5% better-than-random pTRS predictions for neuroscience indications.

**H2:** The use of an LLM-enhanced NLP model leads to a >5% improvement in pTRS prediction over non-LLM for neuroscience indications.

**H3:** A model trained solely on ClinicalTrials.gov data and outcome labels can surpass the predictive performance of industry benchmarks by >5% for neuroscience indications.

**H4:** Use of training data from all indications can improve neuroscience pTRS prediction by >5%.

**1.7 Scope of Research**

The predictive model described here will be trained on all clinical trials in the ClinicalTrials.gov database that existed prior to 2019, and validated on trials from 2019 to June 1, 2024. While some models will be built from trials for all disease indications, attention will be paid to whether training specifically on neuroscience programs increases or decreases the predictive validity of the pTRS estimates for neuroscience programs. Separate models will exist for Phase 1, 2, and 3 trials. This model approach will be applicable for any R&D group looking to improve its pTRS estimates using analysis of their own clinical trial properties, or looking to determine which textual attributes of trial description data most closely correlate with success or failure.

**1.8 Research Limitations**

While the predictive model built in this research will be trained on all data within the ClinicalTrials.gov database within the specified date ranges, there may be trial data from older studies that is missing or incomplete. While BERT is an advanced LLM model that is capable of ingesting and interpreting a variety of language styles, clinical trial data contains medical language that the model may not be optimized for. The model may not be equally applicable across all indications, and it may be better at making predictions for some phases than others, due to the types of information available at the time that the clinical trial description is written. While the model shown here will be able to incorporate the content of a clinical study's design, it cannot directly capture nuanced, undescribed characteristics of the program itself, or the efficacy or safety demonstrated by the program in its previous clinical trials. Conversely, expert elicitation processes for pTRS estimation can easily take these factors into consideration.

**1.9 Organization of Praxis**

This praxis is divided into five chapters. Following this introduction, Chapter 2 surveys the relevant literature regarding pharmaceutical R&D risks, and the various methods employed to develop pTRS estimates. These include both LLM- and non-LLM-based approaches that researchers have utilized to estimate pTRS. Chapter 3 provides an overview of the methodology employed in this research, and in particular, the construction and implementation of the predictive models. Chapter 4 covers results of model validation, comparison with industry benchmarks and other research in this field, and exploration of the most significant factors identified by the model. Chapter 5 summarizes the work, providing potential use cases as well as avenues for future work.

# Chapter 2—Literature Review

**2.1 Introduction**

Publicly traded pharmaceutical companies dedicate more than 25% of their net income to R&D, compared to 2-3% for companies within the S&P 500 as a whole (Austin & Hayford, 2021). This level of investment intensity is needed because of the sheer cost to bring a drug to market, median estimates of which range from $1.14 billion to $2.87 billion (DiMasi, Grabowski, and Hansen, 2016; Wouters & Kesselheim, 2024). These high costs of development require pharmaceutical firms to be as judicious as possible when determining which drug development programs to pursue and which to terminate.

Multiple metrics must be considered when determining the viability of prospective programs. These include tradeoffs between development **costs**, commercial **revenue** projections (with consideration of competitive landscape), **time**/duration of development, and **risk** (Grudzinskas, 2022). Risk is often characterized by examining the *probability of success*, which can include technical success (pTS), regulatory success (pRS) or the probability of both technical and regulatory success (**pTRS**) (Stalder, 2022).

Once a pTRS estimate is available, a company may then value a program asset by risk-adjusting discounted cash flows to generate an expected net present value, or eNPV. Since pTRS dictates the likelihood that a drug candidate molecule will progress forward and be launched, it is a critical parameter in the determination of program value (Harpum, 2010). Improvements in estimates of pTRS allow firms to better identify and reject programs that are unlikely to succeed, and invest in programs that are likely to help patients. (Wong et al., 2019)

Particular attention is also paid to pTRS because of the <u>sensitivity</u> of program value analysis to this variable as an input. As Peck (2017) points out, "relatively small improvements in success rates, especially in expensive clinical development can substantially decrease the total cost of drug development." A sensitivity analysis by Paul, et al. (2010) demonstrates this, showing that success probabilities represent four of the six most sensitive inputs when computing capitalized cost per launch (see Figure 2-1, below).

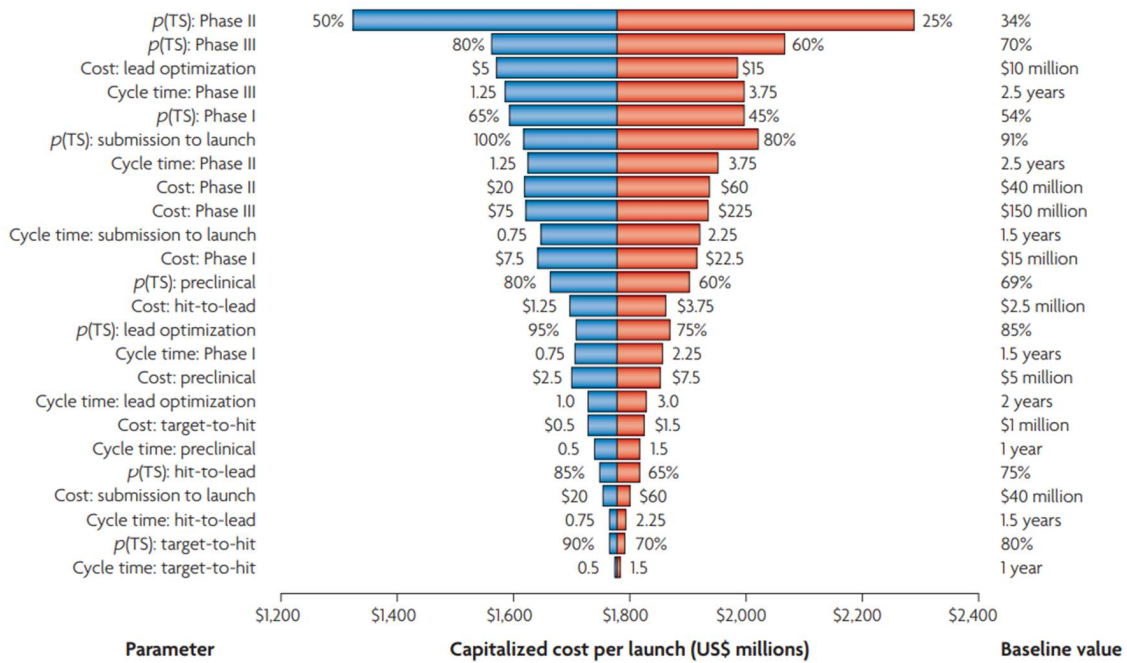

**Figure 2-1. R&D Productivity Model: Parametric Sensitivity Analysis from Paul, et al., 2010. Reproduced with permission.**

Neuroscience is a particularly resource-strained area of drug development. More money is spent for CNS drugs than any other therapeutic area, but only 8% of these drugs even make it to clinical trials, about half the rate of other indication categories (Berger, et al., 2013). Despite enormous unmet need, some prominent CNS disease indications have no viable drug treatments available (Zhu, 2021). Some reasons for this difficulty include the sensitivity of the nervous system to toxicity and the presence of the blood brain barrier, which prevents 95% of potential neuromodulatory molecules from being viable drug candidates (Dong, 2018).

**2.2 Overview of Current Approaches to pTRS Estimation**

Various methods have been developed by which companies may attempt to estimate the probabilities of technical and regulatory success for their products. These include:

1. Industry or internal benchmark pTRS, based on past successes and failures
2. Expert elicitation, which leverages the opinions of knowledgeable individuals
3. Statistical analysis, which can involve many types of information sources
   (Vergetis, 2021)

Each of these methods will now be elaborated upon further.

*Benchmarks*

Pharmaceutical pTRS benchmarks are the literal percentages of programs that succeed in a given phase. These values may be generated deterministically based upon past trial and registration successes and failures, either from the industry as a whole (industry benchmarks), or from a company's internal historical data (an internal or company benchmark). This can be broken down by phase, clinical indication, or both, so long as sufficient data exists to generate meaningful probabilities. (Hampson et al., 2022)

However, benchmarks do not include any specific information or about, or customization for, any one trial or drug candidate. This method is simple and objective, but blind to a drug's clinical development history in previous trials, as well as the intrinsic properties of the individual molecule. Zhou and Johnson (2018) reviewed eight papers from 2010-2018 using different methodologies to generate benchmarks, resulting in overall success estimates ranging from 8.3% to 13.4%.

While many benchmark approaches examine a phase-by-phase approach, which estimates the probability of phase success from a random sampling of phase transitions, Wong, et al. (2019) were able to implement a path-by-path approach that "traces the proportion of development paths that make it from one phase to the next." By composing trials as chains of distinct development pathways and inferring terminations from lack of successive trials, they were able to estimate aggregate success rates for the over 400,000 trials available at the time of publication. While this method does not involve manual curation of study results to develop these aggregates, it does eliminate the selection bias involved in picking a sampling of trials from which an aggregate estimate could be made (the phase-by-phase approach).

*Expert Elicitation*

To build greater sophistication and specificity into the pTRS estimation process, some researchers turned to expert elicitation, which involves recruiting and questioning subject matter experts (SMEs) about the merits of the therapeutics under examination. Eli Lilly and Company was one of the early adopters of this method, and has used an independent review board develop pTRS estimates since 1997. (Andersen, 2012). In 2018, Dallow, et al., of GlaxoSmithKline, implemented a structured Bayesian approach for this known as the SHeffield ELicitation Framework (SHELF) (Dallow et al., 2018). This approach had been originally developed at the University of Sheffield for microbial risk assessment (Oakley & O'Hagan, 2022). The broader approach can be summarized as follows: Experts are selected and trained on the process before being provided with an evidence dossier detailing relevant information about subject at hand. These experts then individually answer questions about their beliefs about one or more uncertain quantities (for example, expected drug safety profile), and responses are used to generate a probability distribution representing the experts' beliefs and uncertainties about those beliefs. SMEs are then allowed to share and discuss their responses, and ideally, come to some group consensus.

  Other firms in the pharmaceutical industry, such as Novartis, now also use a similar SHELF-based approach (Hampson, et al., 2021; Holzhauer, et al., 2022). Benefits of this approach include transparency of calculation, the ability for SMEs to adjust their impression of a drug based on clinical study characteristics, and the ability to leverage the collective depth of expertise within the SME group (Dallow et al., 2018). But the use of human subjects also inherently introduces bias, including over-optimism among opinion leaders, or who may misunderstand the probabilities being asked of them.

  In addition to the SHELF method, another means of deriving expert elicitations include Cooke's method, which aggregates the probability density functions from each expert opinion into a single one, while the Delphi method utilizes an interactive survey approach to enable structured interactions to promote the integration of new information from expert peers (European Food Safety Authority, 2014). However, each of the elicitation methods described here are hampered by the fact that significant time must be siphoned from busy, valuable members of organizations to enable these interactions. That time investment, as well as the biases of experts, have led many to look for faster, more objective means of pTRS estimation.

*Statistical Methods*

Statistical methods for PTRS estimation can take on many forms. For purposes of this review, these methods will include any attempt to use objective data to improve pTRS estimation, beyond what would be provided by a simple benchmark. These tend to fall into two categories, big data approaches (examining large datasets for attributes that correlate with success) and mechanistic approaches, which use the biophysics of the drug's interaction with biological systems to predict efficacy, toxicity, etc.

Attribute analysis of the drug molecule and/or target are correlated to the success of drugs of the past to generate predictions of pTRS. Yamaguchi, et al. (2020) focused on **drug target** (enzyme, receptor, etc.), **action** (agonist, antagonist, etc.) **modality** (monoclonal antibody, small molecule, etc.), and **application** (cardiovascular, nervous system, etc.) for 3999 molecules. They were able to determine success rates within each category, enabling the construction of multivariate predictive models of pTRS based on these factors. These factors, and others, can then be used to generated more specific, tailored industry benchmarks (Hampson, et al., 2022). Big data analysis such as this has the ability to leverage decades of real-world clinical results, but is sensitive to incorrect or misattributed information within that data.

With the advancement of artificial intelligence and *in-silico* quantitative pharmacology approaches, it is also possible to directly interrogate and predict the efficacy and off-target effects of novel compounds based on molecular physics (VeriSIM

Life, 2024). Some evidence exists that this may give rise to improved success predictions. These mechanistic approaches do not rely on existing (potentially flawed) datasets, and are more generalizable in that can make predictions for highly novel interventional approaches (Geerts, et al., 2018).

**2.3 NLP Models and LLMs**

*NLP Models*

Recent developments in natural language processing (NLP) and large language models (LLM) have expanded the capability to ingest and utilize large text-based data sets. NLP can be divided into the areas of NLU (natural language understanding) and NLG (natural language generation) (Khurana et al., 2023). Highly functional NLU requires that a system be able to extract the morphology, syntax, semantics, and context of textual data up to a sentence long. Beyond one sentence in length is known as discourse, the level at which NLP deals with longer segments of text.

A significant problem in NLP is known as the "curse of dimensionality" – the fact that the sequence of tokens upon which a model will be tested will usually not be included in its training set. (Pappas & Meyer, 2012). Neural language modeling (NLM) was proposed in the early 2000s as a way to predict a succeeding word based on a probabilistic model taking into account the previous n words (Bengio et al., 2003). Despite these levels at which NLP can break down text to constituent elements to extract meaning, ambiguity of meaning can still exist, a problem tackled by Gao and others in the 2010s (Gao et al., 2015; Umber & Bajwa, 2011).

*LLMs*

In recent years, large language models (LLMs) have taken on a rapidly expanding role in the NLP ecosystem. According to Naveed, et al. (2024), the number of published papers containing the term "large language model" has increased from only 60 in 2019 to over 28,000 as of 2024.

Much of this progress has been related to the development of transformers, underlying algorithms that power many modern LLMs (Luitse & Denkena, 2021; Vaswani et al., 2017). Raiaan et al., (2024) state that, "The transformer model has had a significant impact on the field of NLP and has played a crucial role in the development of language models such as Bidirectional Encoder Representations from Transformers (BERT)." The pre-trained BERT model can easily be tuned to accomplish a range of tasks without requiring changes to its core architecture (Devlin, et al., 2019). The bidirectionality of this transformer-based encoder means that it can take into account tokens both before and after a word in order to account for semantics (Li et al., 2022). LLMs have also proven to be useful tools for Biomedical Natural Language Processing (BioNLP) applications, which require the ability to process nuanced and often obscure medical text (Chen et al., 2023). Because of their power and ease of deployment, the BERT family of LLMs will be used for the work described in this praxis.

**2.4 AI-based Methods for Clinical Trials**

Pharmaceutical firms employ a range of technical functions to execute the design and development of clinical trials, a task which implicitly requires examination of steps to improve the risk profile of clinical trials for candidate drugs. These areas include Clinical Development and Operations, Regulatory Affairs, Medical Affairs, Biostatistics, and Program & Portfolio Analytics. Each of these groups can influence clinical trial design, and thus, clinical risk profile, and artificial intelligence has shown promise in delivering value for many of these functions.

**Table 2-1: Areas of Machine Learning Contribution to Clinical Research. Adapted from Weissler, et al. (2021).**

| Clinical Trials and Observational Research | | |
|---|---|---|
| Pretrial Planning | Participant Management | Data Management |
| Protocol Development | Cohort Selection | Automate Data Collection |
| Drug Regimen Selection | Patient Identification | Monitor Data Quality |
| Site Selection | Participant Retention | Adjudicate Outcome Events |
| | | Analyze large, highly dimensional, or sparse datasets |
| | | Unlock novel biological features |

As shown in Table 2-1, above, there are a number of ways in which machine learning can aid the clinical development process. In particular, significant strides have been made in using ML to aid in the design and execution of clinical trials (Kolluri et al., 2022). Li, et al. (2015) developed a dose-response framework that could predict clinically-relevant translational biomarkers, a critical asset that development teams need to demonstrate efficacy. Li, et al. (2020) and Liu, et al. (2020) were able to use Bayesian learning to construct models to aid design of both adaptive and modified toxicity probability interval (mTPI) dose finding studies, respectively.

Strides have also been made in clinical trial execution, a critical step in getting therapies to market in a timely manner. Beck, et al. (2020) developed an AI-based tool to help determine patient eligibility, an otherwise cumbersome and manual process, by screening patient medical records. In 2021, Haddad, et al., evaluated an AI-based CDSS (clinical decision support system) and found an overall accuracy of 87.6% in screening patients for a cancer trial. Analysis of imaging also plays a role, as researchers were recently able to develop AI-based tools to identify patients potentially eligible for clinical trials based on their OCT (optical coherence tomography) retinal scans (Williamson, et al., 2024).

**2.5 AI-based Methods of pTRS Estimation**

This praxis approaches the problem of pTRS estimation by leveraging large datasets, employing natural language processing. This section will describe the work that has been done in this area, and the existing gaps.

Approaches to estimating pTRS can be developed either as chemical component models (for example, predicting efficacy or safety based on molecular structure), or as composite models that incorporate text and other elements to predict pTRS directly.

*Component Models and Techniques*

Several estimable properties of molecules themselves are known significantly affect pTRS, especially safety and efficacy. Gayvert et al. (2016) developed an approach called PrOCTOR that uses a compound's structure and molecular targets, identifies properties that could give rise to safety concerns, and used this information to the probability of such safety concerns preventing drug candidate success. Their model achieved an ROC (receiver operator curve) AUC of 0.826.

Similar approaches have been taken on the efficacy side of the analysis. Computational approaches to understanding structure-activity relationships have been going on for decades and have greatly accelerated as a result of recent developments in artificial intelligence. In 2015, Wallach, et al., developed AtomNet, a deep convolutional neural network to predict biological activity of drug molecules. Murali, et al., (2022) used an ML-based approach to predict pTRS based on predicted biological activity of drug candidates. Other work further developed these techniques, which are now used as tools to help design drugs with improved pTRS (Staszak et al., 2022). Approaches such as these are also used in many composite pTRS prediction tools, which use structural data along with other inputs to inform pTRS estimates.

*Text-Based and Composite pTRS Models and Techniques*

Composite models are built to generate an overall prediction of success for a candidate drug, be it for a single trial or phase, or its overall development. These approaches may utilize clinical trial descriptions, trial design parameters, and/or structural/molecular data, to generate predictions for success. Such models that utilize AI/ML/NLP will be described here.

A significant foray into this approach was accomplished by Lo, et al. (2017), using 6344 drugs, derived from the databases Pharmaprojects and Trialtrove. 31 drug attributes and 113 clinical trial characteristics were used as features for ML-based prediction using R. Their analysis resulted in a ROC-AUC of 0.78 for Phase 2 and 0.81 for Phase 3 trials. While the data was limited in time frame (1990-2015) and excluded Phase 1 prediction, the ability of this algorithm to outperform complete-case analysis was noteworthy.

Subsequent ML-based approaches utilized different combinations of source data and core methodology. Feijoo, et al. (2020) used a combination of ClinicalTrials.gov and Biomedtracker data to generate a dataset of 6417 individual trials that initiated between 1993 and 2004. A supervised random forest model was implemented, which led to a reported all-disease accuracy, sensitivity, and specificity of (0.743, 0.740, 0.746) for Phase 2 trials and (0.672, 0.692, 0.648) for Phase 3 trials.

Fu, et al. (2022) introduced HINT (Hierarchical Interaction Network) as a graph neural network method for clinical trial outcome prediction. The method uses an interaction graph to connect knowledge-embedding modules containing information on the drug, disease, etc. 9045 total trials were used for training and validation, and tested on 3420 total trials, achieving ROC-AUC of (.576, .645, and .723) on Phases 1, 2, and 3, respectively. Updated publications on this method are available from Lu, et al. (2024).

Meanwhile, Aliper, et al. (2023) developed inClinico, a multi-modal AI prediction platform designed specifically to predict Phase 2->Phase 3 transitions. 55,653 unique Phase 2 clinical trials were used in the proprietary training set and 2849 trials were used for validation, achieving a robust ROC-AUC of 0.88 and 79% accuracy for real-world trials that have read out.

Reinisch, et al. (2024) created CTP-LLM, a GPT 3.5-based model designed to predict clinical phase transitions. Using a combination of ClinicalTrials.gov and Biomedtracker data, they obtained a training data set of 20,000 trials and based on published trial protocols generated five models, with their best model achieving a ROC-AUC of 0.667 and an F1 score of 0.665.

Zheng, et al. (2024) also used a multimodal LLM approach they dubbed LIFTED as an approach for clinical trial outcome prediction. Their mixture-of-experts approach transforms modality data into natural language descriptions for LLM models. The mixture-of-experts framework can then identify common information patterns across modalities. The composite LIFTED method was able to achieve ROC-AUC scores of (.649, .651, .735) for Phases 1, 2, and 3.

A dissertation by Jung Won Choi in (2024) laid out a comprehensive approach to the use of machine learning for clinical trial analysis, and included some outcome prediction. Choi used a random forest model in R to make predictions based on trials in which *only a single country* was used for trial sites (n = 83,852). Across all phases, a mean ROC-AUC of 0.85 and mean F1 of 0.95 was achieved.

In order to address the challenges of prediction on non-small molecule entities as well as the problems inherent to graph neural network approaches, Gao et al. (2024) introduced LINT (Language Interaction Network), a method requiring only clinical trial text input. Descriptions, pharmacodynamic and pharmacokinetic data were retrieved from DrugBank, and LINT uses a classifier and transformer to generate a learning model. Despite tackling only hard-to-predict biologics candidates, the model achieved ROC-AUC scores of (.770, .740, .748) in Phases 1, 2, and 3, respectively.

**Table 2-2. Comparison of Phase 2 Predictive Performance Across Models in Literature**

| Paper | N | Scope of Training Data | Performance |
|---|---|---|---|
| Lo, et al. (2017) | 6344 drugs | Trial characteristics + drug attributes | ROC-AUC: Ph2 = **0.78**; Ph3 = 0.81 |
| Feijoo, et al. (2020) | 6417 trials | Trial characteristics + Biomedtracker data | ROC-AUC: Ph2 = **0.76** for Neuro |
| Gao, et al. (2024) | 23,519 trials | Trial characteristics + drug attributes | ROC-AUC: Ph2 = **0.74** |
| Fu, et al. (2022) | 9045 trials | Trial characteristics, some drug attributes | ROC-AUC: Ph2 = **0.65** |
| Aliper, et al. (2023) | 55,653 trials | Multimodal | ROC-AUC: Ph2 = **0.88** |
| Reinisch, et al. (2024) | 20,000 trials | Trial characteristics + Biomedtracker data | Ph2 Accuracy = 0.604; F1 = 0.600 |
| Zheng, et al. (2024) | 17,538 trials | Trial characteristics + drug attributes | ROC-AUC: Ph2 = **0.651** |

Gao, et al. (2024) also released a paper introducing CTO – the Clinical Trial Outcomes dataset. Much of the limitation in generating large training sets using the complete index of trials on ClinicalTrials.gov came from the lack of a single source of truth regarding the actual success/failure outcomes of those trials. Manual curation by some companies has yielded proprietary datasets comprising only a small (and possibly nonrepresentative) fraction of those trials. Gao and colleagues were able to develop a machine learning model using publications, news reports, trial linking, etc., to generate a dataset that predicts with remarkable accuracy whether a completed trial was successful or not. The random forest model they created has a 91% overall accuracy, sufficient for use in validating clinical trial predictive models. Most importantly, its predictions are available for nearly all interventional trials contained within ClinicalTrials.gov, representing an enormous leap in scope for predictive validation. The CTO Database will be used as the source of truth for trial successes and failures in model validation for this praxis.

## 2.6 Prediction Validation Methods

Trained NLP-based models require formal evaluation before they can be trusted for implementation. Because different metrics focus on different aspects of performance, several should be used in order to understand the strengths and weaknesses of a model, understand its performance relative to existing models, and identify areas for improvement.

This praxis concerns itself with the estimation of the probability of success for program phases. This could be validated using binary predictions of clinical trial successes and failures and evaluated relative to the true outcomes. However, it may be more prudent for the model to generate a pTRS value for each clinical trial and evaluate those continuous values relative to the binary outcomes. Both general approaches to validation will be described here.

### *Binary Prediction Validation Methods*

Binary classification results in binary predictions, and is often used when an outcome is all-or-nothing, such as detecting disease, patient survival, etc. The true state of whether an event occurs is classified as positive or negative, while the prediction may be classified as true or false.

The most commonly applied metrics for binary classification models are accuracy, precision, recall, and F1 score (Naidu et al., 2023). These are derived from the values contained within the confusion matrix of true positive (TP), true negative (TN), false positive (FP) and false negative (FN) (Hossain, et al., 2023).

Accuracy is the proportion of all predictions that were true (Equation 1):

$$Accuracy = \frac{TP+TN}{TP+TN+FP+FN} \quad (1)$$

Precision is the proportion of all positive predictions that were true (Equation 2):

$$Precision = \frac{TP}{TP+FP} \quad (2)$$

Recall is the proportion of true predictions out of all of the positive events (Equation 3):

$$Recall = \frac{TP}{TP+} \quad (3)$$

F1 Score is the harmonic mean of Precision and Recall (Equation 4; Hand et al., 2021):

$$F1 = 2 * \frac{Precision*Recall}{Precision+Recall} \quad (4)$$

Balanced accuracy (sometimes referred to as the "index of balanced accuracy") is another means by which model and prediction accuracy can be gauged, which can account for imbalances in data. A balanced accuracy score of 0.50 represents that a model is no better than random guessing. (García et al., 2009)

Another useful method in model development is K-fold cross validation, which can prevent overfitting. This technique requires that a model be tested on *k* different subsets, or folds, of the test data, and trained on the rest (Wilimitis & Walsh, 2023). By preventing overfitting, the researcher may reduce the chance of a model performing well on training data, but being unable to make accurate predictions when presented with new data (Hossain, et al., 2023). K-fold cross validation may be useful for either binary or continuous classification schemes.

*Continuous prediction validation methods*

Receiver operating characteristic (ROC) curves plot true positive rate against false positive rate across a series of thresholds (Nahm, 2022). The area under the curve (ROC-AUC) can then be measured as a way of characterizing the relative predictive performance of the model.

Precision-Recall curves are another tool for understanding the interplay between various elements of the confusion matrix. Specifically, PR-AUC can be very valuable when there is a stark imbalance in data, for example if positive events are rare, for example in fraud detection (Cook & Ramadas, 2020). Because the outcome data in this praxis is imbalanced, PR-AUC will be examined.

Cohen's kappa is a metric that can be used to measure agreement between two sets of predictions, but is able to account for prediction agreement that would result from random chance (Grandini et al., 2020).

Brier Score measures the mean squared error between predictions and outcomes (Zhu, et al., 2024). It is an overall accuracy metric, and thus can evaluate both discrimination (discerning the positive and negative) and calibration (probabilities vs. actual event frequencies).

Log Loss Score attempts to measure the distance between a predicted distribution and its true distribution, and penalizes incorrect classifications (Aggarwal et al., 2020).

**2.7 Feature Importance Analysis**

A common challenge with machine learning models is understanding why an accurate model is making the predictions that it makes. Tools and techniques are available to allow NLP models to be more transparent, by lending insight into the characteristics that lead to certain categories of predictions.

In the work described in Chapter 3, a technique called SHAP (SHapely Additive exPlanations) is implemented to improve understanding of the reasons for model predictions. This allows for the identification of features that most strongly impact the model's predictions, as well as a measure of the magnitude and direction of that influence, increasing or decreasing a predicted pTRS, in the case of this work (Mosca et al., 2022).

**2.8 Summary and Conclusion**

Pharmaceutical research and development involves enormous capital expenditure and risk, so appropriate identification of a program's likelihood of achieving technical and regulatory success is of paramount importance. Not only does pursuing low-viability drug candidates destroy value and expose trial participants to potentially toxic or non-efficacious treatments, there is also considerable opportunity cost, as companies must forego other potentially promising therapies. Because of this importance, efforts for predicting trial outcomes has been of keen interest in both industry and academia.

Approaches that firms currently employ range from having no formal system of estimating pTRS at all, to using benchmarks alone, benchmarks predicated on indication or other factors, expert elicitation to develop more bespoke risk profiles, or any combination thereof. Each of these approaches involves a different level of time, effort, and potential reward in terms of accuracy.

The evolution of machine learning, natural language processing, and large language models have over time provided a variety of new tools with potential for improving clinical development. Some of this includes machine learning based applications to help researchers recruit and select appropriate patients for clinical trials, or identify the most appropriate biomarkers for clinical analysis.

Similar tools have been developed to analyze drug candidates themselves. Numerous attempts have been made to develop means for predicting the toxicological profile of new drugs, based on their structure, targets, and known databases of existing drugs. Similar attempts have been made to estimate potential efficacy. While these can inform some preclinical and clinical decision making, they do not serve, by themselves, to quantify pTRS for a drug candidate.

The latest generation of AI-based methods for pTRS estimation take advantage of tools that can help extract insights from enormous amounts of data. Some of these methods are largely numerically-based and utilize feature extraction to define estimates based on any number of properties of a trial or molecule, from patient count, to binding site characteristics. Other, more straightforward tools may rely solely upon text and utilize NLP/LLM models to derive predictions from subscription or publicly available trial data.

A major limiting factor has been the fact that trials listed in major comprehensive databases such as ClinicalTrials.gov do not contain information on whether the trial was a "success" or "failure." Efforts for manual curation of these trials has resulted in smaller (<20,000 study) databases for training and validating predictive models. However, there is opportunity for bias, as the trials for which outcomes have been curated may be more specific to a particular time, region, indication, or other property that could skew results. Recent work by Gao, et al., to generate robust success assessments for completed clinical trials has afforded success labeling for trials in model training sets. This has made it possible to utilize the full canon of clinical information on ClinicalTrials.gov, enabling more powerful LLM-based development of pTRS prediction models based solely on publicly available information.

# Chapter 3—Methodology

## 3.1 Introduction

This work approaches the problem of pTRS similarly to many of the papers detailed in Section 2.6 – deployment of machine learning as a tool to improve predictions. Openly-available ClinicalTrials.gov data for 101,145 clinical trials was used to train and test the model, using the success predictions of Gao, et al. (2024) as the source of truth, which enabled use of the entire canon of past trials rather than a small curated fraction. This much larger starting dataset provided a sufficiently large set of neurology programs for neurology-specific NLP training.

This chapter will provide the step-by-step overview of the methods implemented to develop and validate these predictions. Section 3.2 provides information on data collection and format. Section 3.3 provides the preprocessing framework that was employed to prepare data for analysis. Section 3.4 details the non-LLM models. Section 3.5 details the screening of the different LLM-enabled models. Section 3.6 provides information on the BioBERT LLM-based final model. Finally, Section 3.7 provides a general description of appropriate implementation.

## 3.2 Data Collection

There were three primary data components used for this analysis. The ClinicalTrials.gov data provided the trial information from which the model was built. The Clinical Trial Outcomes Dataset from Gao, et al. (2024) provided the source of truth for the trial outcomes. Finally, clinical benchmarks for overall and neurological study indications were used as a comparison to determine whether the predictive model was superior to non-program-specific, naïve benchmarking.

*Clinical Trial Data*

Clinical Trial Data was downloaded directly from ClinicalTrials.gov as a CSV file. From 525,418 initial database entries, a downloadable set of 235,200 after filtering for "Interventional" trials, selecting for only those labeled "Completed" or "Terminated", and selecting only those for which the study completion date was earlier than 1 April 2024. The date selection was to remove those trials for which the Clinical Trial Outcomes database would not have a trial outcome prediction.

The download itself allows for selection of individual columns that the user wants for the CSV. Columns believed to be irrelevant were unselected to remove noise and size from the data. The removed columns included "Study URL", "Acronym", "Study Results", "Other IDs", "First Posted", "Results First Posted", "Last Update Posted", and "Study Documents". Remaining columns included "NCT Number", "Study Title", "Study Status", "Brief Summary", "Conditions", "Interventions", "Primary Outcome Measures", "Secondary Outcome Measures", "Other Outcome Measures", "Sponsor", "Collaborators", "Sex", "Age", "Phases", "Enrollment", "Funder Type", "Study Type", "Study Design", "Start Date", "Primary Completion Date", "Completion Date", and "Locations." Note that the model was not trained on "Study Status", "Start Date", "Primary Completion Date", or "Completion Date", as these were used for sorting and preprocessing but omitted prior to training.

*Clinical Trial Outcomes Database (CTOD) Data*

The Gao paper (2024) provides links to the CTOD data used as the source of truth for clinical trial outcomes. The CSV files containing this data are differentiated by Phase number, and contain the NCT Number (a clinical trial ID number) for each trial evaluated, along with the predictions, both binary and continuous.

*Benchmark Data*

Two benchmarks were generated for model evaluation. The first was data from Citeline's Pharmapremia Pharma Intelligence (n.d.) database, which provides current information on clinical trial success rates from 2015-2024. This was used as an industry benchmark for Phases 1, 2 and 3. The second was the calculated success rate for Phases 1-3 in the overall CTOD dataset itself (see Section 3.6 for additional details).

**3.3 Data Preprocessing**

A number of steps were undertaken to address missing data, merge predictions with the clinical data, reformat entries, do basic duration calculations, properly assign phases, and eliminate trials for which key pieces of information were missing.

*Merging Data*

A python script was written to ingest the CTO dataset files (one for each of Phase 1, Phase 2, and Phase 3) and the exported ClinicalTrials.gov ("CT.gov") dataset. Since both files differentiate trials by NCT Number, the script merges the predictions with the rows containing the trial data, and adds new cells within the CSV containing the clinical trial data.

## Date Processing

*Date Reformatting*

Dates were not supplied in a uniform format across the CT.gov dataset, so a Python script was used to convert the dates to YYYY-MM-DD format, regardless of the input format. If only a month and year were supplied, the date was assumed as the 1$^{st}$.

## Phase Formatting

CT.gov data comes with phase labels for each row/trial. These include "EARLY_PHASE1", "PHASE1", "PHASE1|PHASE2", "PHASE2", "PHASE2|PHASE3", "PHASE3", AND "PHASE4". In keeping with the sorting rules used by Gao in the CTO dataset, "EARLY_PHASE1", "PHASE1", and "PHASE1|PHASE2" were labeled as "PHASE1", "PHASE2|PHASE3" was labeled "PHASE2", and all trials labeled "PHASE4" were eliminated.

These phase assignments were selected in order to sort the trials as accurately as possible. Since Phase 1/2 trials still have not discharged Phase 1 risks, they are treated as Phase 1 trials still for purposes of pTRS estimation, and likewise with Phase 2/3 trials being treated as Phase 2. Phase 4 trials are post-approval trials and thus are outside the scope of typical pTRS estimation analysis.

## Purging Incomplete Data

Rows that were missing data for phase, prediction, or completion date were removed from the dataset. After all steps in data preprocessing were completed, 101,145 clinical trials remained, spread across the three phases of interest.

*Filtering for Neuroscience Trials*

ClinicalTrials.gov data does not designate the therapeutic area (e.g., oncology, dermatology, etc.) for a given clinical trial. However, in industry, pTRS values are differentiated not just by phase but also by therapeutic area, and the area of interest for this praxis is neuroscience. This work therefore required that a standalone neuroscience training/test set be constructed.

A list of indications and disease areas in neuroscience developed through use of various databases (see *Appendix B*). The decision was made to be as comprehensive as possible, intending to capture all of the possible neuroscience programs while acknowledging the possibility that some non-neuroscience indication trials might leak through the filter. This allows the neuroscience model to train on rare neurological conditions, and ensures that the resulting train/test set is large enough to facilitate appropriate model validation.

Some indications may be named a number of different ways (e.g., 'Alzheimers Disease', 'Alzheimer's Disease', AD, etc.). To remedy this, ChatGPT-4o was used to expand upon the list and amplify the indication terms by including plural and abbreviated names for each, as well as synonyms. Then, a Python scripts was written that filtered clinical trials for which the "Conditions" data field contained any of the terms. In total, 32,106 trials met the inclusion criteria for the neuroscience training/test sets, across all 3 phases.

*Overall Data Flowchart*

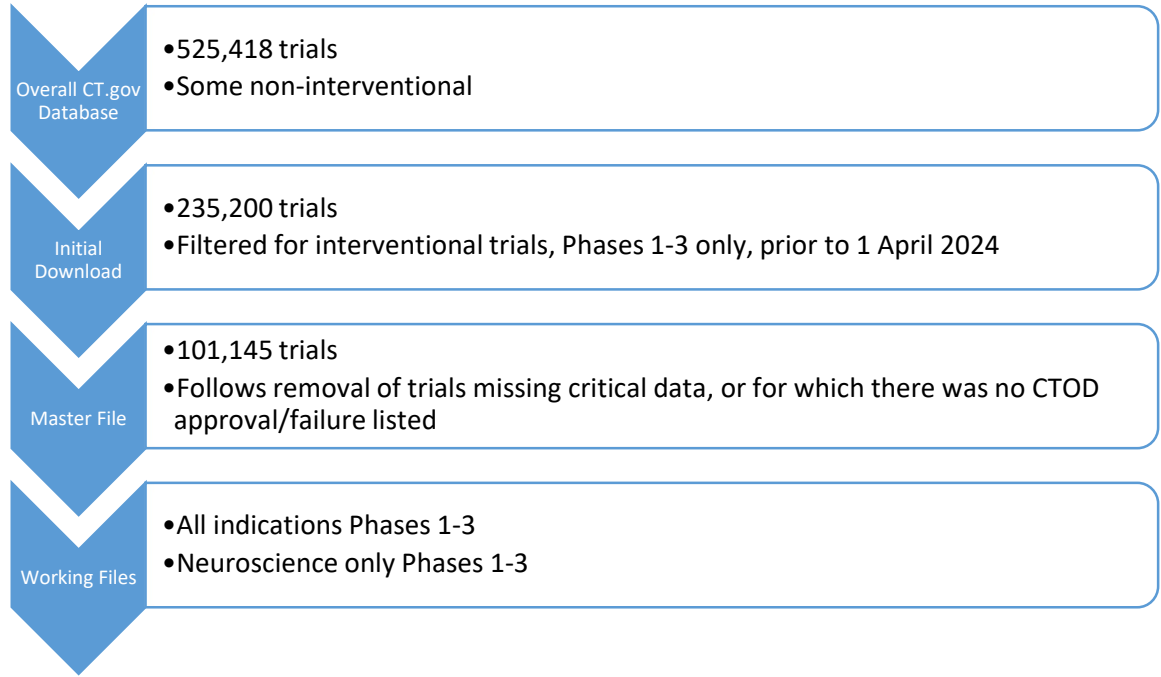

## 3.4 Basic (non-LLM) NLP Models

Three non-LLM NLP models were constructed in order to understand whether accurate inferences about pTRS could be made based on a relatively simple model without the use of sophisticated LLMs. These relatively simple Python models were operated from the PyCharm IDE console, using an ASUS Vivobook laptop PC with an Intel Core i5-1235U processor and 40 GB of RAM.

*Logistic Regression Model Description*

A Python program was written to generate a logistic regression model. The program read in the CSV file of interest, trained on clinical trials that completed before 01 January 2019 and testing on those that completed on or after that date, with the binary success predictions from the CTO Database. The LogisticRegression function within the Scikit-learn machine learning library was used, and metrics included a ROC curve, ROC-AUC, confusion matrix, accuracy, precision, recall, F1, balanced accuracy score, average precision score, and Cohen's kappa. A pilot evaluation run was made for each phase in order to generate an optimal decision threshold value for binary prediction (done by maximizing Youden's J statistic). Then each train/test set (all indications vs. neuroscience only, for each of the three phases) was trained and evaluated under that decision threshold.

*Random Forest Model Description*

A Python program was written to generate a logistic random forest model. The program read in the CSV file of interest, trained on clinical trials that completed before 01 January 2019 and testing on those that completed on or after that date, with the binary success predictions from the CTO Database. The RandomForestClassifier function within the Scikit-learn machine learning library was used, and metrics included a ROC curve, ROC-AUC, confusion matrix, accuracy, precision, recall, F1, balanced accuracy score, average precision score, and Cohen's kappa. A pilot evaluation run was made for each phase in order to generate an optimal decision threshold value for binary prediction (done by maximizing Youden's J statistic). Then each train/test set (all indications vs. neuroscience only, for each of the three phases) was trained and evaluated under that decision threshold.

*Gradient Boosting Model Description*

A Python program was written to generate a gradient boosting model. The program read in the CSV file of interest, trained on clinical trials that completed before 01 January 2019 and testing on those that completed on or after that date, with the binary success predictions from the CTO Database. The GradientBoostingClassifier function within the Scikit-learn machine learning library was used, and metrics included a ROC curve, ROC-AUC, confusion matrix, accuracy, precision, recall, F1, balanced accuracy score, average precision score, and Cohen's kappa. A pilot evaluation run was made for each phase in order to generate an optimal decision threshold value for binary prediction (done by maximizing Youden's J statistic). Then each train/test set (all indications vs. neuroscience only, for each of the three phases) was trained and evaluated under that decision threshold.

*Feature Importance Analysis*

In order to gain a better understanding of the individual training set features that the model considers most important, it was necessary to carry out feature importance analysis. This was done using the gradient boosting model since it was the best performing non-LLM model, and because for LLMs in general, it is considerably more difficult to understand the rationale underlying a given model.

For this work the 20 most important features in the gradient boosting model were extracted and SHAP values were generated to indicate the direction of the influence (i.e., whether a given token increases or decreases predicted probability of success.

*Pseudocode for the Gradient Boosting Model*

The script loads the training and test datasets from CSV files and performs data preprocessing to ensure that all rows and contain valid entries. It filters out rows that are missing the target value ("pred_proba") and invalid dates, then splits the datasets using a cutoff date (pre-2019 for training, 2019 and onward for testing). All text columns (excluding "pred_proba" and the date column) are combined into a single "combined_text" field and renamed from "pred_proba" to "label," converting it to a float while also producing a binary version (by rounding) for metrics like accuracy that require classification. Next, a TF-IDF vectorizer is used to transform the combined text into numerical features, which are then used along with the binary labels to train a gradient boosting classifier.

After training, the script evaluates model performance on the test set, computing various metrics such as ROC-AUC, accuracy, precision, recall, F1 score, and confusion matrix. It then extracts built-in feature importances from the gradient boosting model and logs the top 20 features. In addition, a SHAP analysis is performed: the test set features are converted to a dense float array, and SHAP's TreeExplainer computes the SHAP values, which provide a directional/signed contribution for each feature. The script logs the top 20 features by mean SHAP value (indicating whether they tend to increase or decrease the predicted value). Finally, it computes the ROC curve from the predictions, compares the model's performance with benchmark predictions, calculates the proportion of test cases where the model outperforms these benchmarks, and writes all logs to an output.

**3.5 LLM-Enabled NLP Model Screening**

In order to generate models with greater accuracy, several LLMs from the BERT (Bidirectional Encoder Representations from Transformers) family were evaluated. These included BERT, BioBERT, SciBERT, and PubMedBERT. A single train/test set was chosen (Phase 1, Neuroscience-only) and run for only three training epochs due to the computational intensiveness of training all models on all data sets.

**3.6 BioBERT Models**

Once the BioBERT model had been selected, additional steps were taken to enhance the predictive power of the model. Additional data became available within the CTO dataset that included continuous predictions of success for those trials where it was not definitively clear whether a clinical trial had succeeded or failed. Moving forward, all model training and evaluation was done using the prediction dataset that included continuous values, to help produce more nuanced distinguishment of auspicious trials.

In order to properly validate these continuous predictions, additional metrics were incorporated into the evaluation process, specifically log loss and Brier score. These are useful in evaluating comparisons between continuous predictions and true values. The original binary comparison metrics relating to confusion matrix elements (e.g., ROC-AUC, balanced accuracy, F1, etc.) remain.

There is also value in understanding not just the overall relative accuracy of a pTRS prediction model, but also the proportion of the time that a model prediction is superior to a benchmark. This analysis was included as well.

*Benchmark Comparisons*

*Industry Benchmarks*

Two benchmarks were generated for this project. The industry benchmark is a weighted average of the industry benchmarks for success rates in phases 1, 2, and 3, for psychiatry and neurology These were weighted by number of studies used to make the benchmark (the constituent trials were 76% neurology and 24% psychiatry) to generate a benchmark for neuroscience as a whole. Performance superior to this benchmark indicates the model predictions are better than simply assigning the published industry benchmarks for each study in each phase of the test set. These benchmarks were compiled by Citeline, Inc., and represent the alternative pTRS that a company would have to calibrate their asset valuations if they relied on benchmarks alone. Predictive ability superior to the industry benchmark would be considered **a minimum requirement** for a pTRS prediction model to be viable.

**Table 3-1. Industry Benchmarks (derived from Citeline data)**

| Phase 1 | Phase 2 | Phase 3 |
|---------|---------|---------|
| 48.2%   | 29.0%   | 50.5%   |

*Dataset-specific Benchmarks*

Dataset-specific benchmarks were also generated. This benchmark represents the percentage of successes predicted in the CTO Dataset used for success labeling. For each phase in the neuroscience-filtered test sets, the 'pred_proba' column was averaged to generate an average success rate across the specific dataset being used. Performance superior to this benchmark indicates the model predictions are better than simply assigning the dataset's mean phase success probability to each trial. It should be noted that the dataset-specific benchmark also included the outcomes for studies in the test set, which the model is not trained on, making it significantly more challenging to beat, since the benchmark is predicated on outcome data that is not used to inform the model. This is a less realistic comparator than the industry benchmark, since this information would not be available to a firm at the time it is developing program analysis and making decisions on pursuing drug candidates. However, this benchmark was used to further characterize the predictive ability of the model. Predictive ability superior to the dataset benchmark would be considered **aspirational** for a pTRS prediction model.

**Table 3-2. Dataset-Specific Benchmarks (derived from the predictions of success in the CTO Dataset)**

| Phase 1 | Phase 2 | Phase 3 |
|---|---|---|
| 63.2% | 56.7% | 65.4% |

*Epoch Control*

In order to prevent over or underfitting, it is important to control the number of training epochs in each model. This was accomplished with respect to a variable called training loss. The training script continued to initiate consecutive training epochs until the training loss had failed to decrease for three consecutive training epochs. At that time, training stops and the model is evaluated.

*BioBERT Code Description*

The script first loads training and test data from CSV files and filters them based on the completion dates (using trials before 2019 for training and trials from 2019 onward for testing). To ensure that no invalid entries remain from initial preprocessing, the files are preprocessed once again. Samples missing the target column ("pred_proba") are dropped, and dates are converted to proper date types. Next, the script combines all text columns (except the target and date fields) into a single "combined_text" string for each sample. The "pred_proba" column is then renamed to "label" and converted to float (with a binary label version created via rounding for metrics such as accuracy that require binary values).

After preprocessing, the script initializes a BioBERT model by loading the "dmis-lab/biobert-base-cased-v1.1" checkpoint. It trains the model using a training loop that minimizes binary cross-entropy loss with early stopping (if the loss does not improve for a three epochs). After training the model, the script evaluates the model on both the training and test sets. Various metrics (ROC-AUC, balanced accuracy, precision, recall, F1 score, etc.) are computed by rounding the continuous labels (except for the final models in sections 4.5-4.7, which were trained using the continuous labels). The script also compares the model's predictions to two fixed benchmark values (an "industry benchmark" and a "dataset benchmark") by creating constant prediction arrays and computing evaluation metrics as well as the proportion of cases where the model's prediction is closer to the true value than the benchmark's. Finally, the script outputs the ROC curve as an image file and logs all the evaluation details to a text file.

**3.7 Implementation**

In order to implement the model described in this praxis, the user will need to ascertain some basic information about the clinical trial(s) in question. For example, if a user intends to use the model to predict the probability of Phase 2 success for a program, they will need to implement the model based on the trial(s) that comprise Phase 2 for that program.

The model described here is trained on the majority of clinical trial data that accompanies descriptions in ClinicalTrials.gov datasets. Therefore, inputs to any implementation will include those types of data, which include the following: "NCT Number", "Study Title", "Brief Summary", "Conditions", "Interventions", "Primary Outcome Measures", "Secondary Outcome Measures", "Other Outcome Measures", "Sponsor", "Collaborators", "Sex", "Age", "Phases", "Enrollment", "Funder Type", "Study Type", "Study Design", and "Locations."

For trials that have already begun enrolling patients, or which are registered in ClinicalTrials.gov, acquiring these inputs will not be a particularly difficult. For those for which this information is not yet available, it may be quite difficult to use this model for predictions. Since the model relies on properties of the clinical trial itself, generating surrogate inputs for these fields, for a clinical trial that has yet to be designed, would likely not result in quality predictions. Once all inputs for a single trial are available, they should be loaded into a single row of a CSV file and used as an input for the model, with code implemented to output the prediction to console or append to an unused column in the row.

Practical applications that drug development professionals will have for this model fall into two primary areas:

1. Analysis of internal programs in development

2. Analysis of external competitor programs

The former is typically the business of internal finance and portfolio management groups, while the latter tends to be the domain of commercial development experts who wish to understand an internal asset's chances of being first-in-class, an early follower, a late follower, etc., as this can have ramifications for revenue projections. Regardless of application, however, the above-described inputs will be needed in order to generate a high quality pTRS estimate.

## Chapter 4—Results

### 4.1 Introduction

The datasets were preprocessed, and the code scripts were prepared in Python as described in Chapter 3. The metrics of interest were then compiled and visualized using Excel®. This chapter will describe the results from the following:

- Non-LLM models
    - Logistic Regression
    - Gradient Boosting
        - Feature extraction/SHAP
    - Random Forest
- LLM model comparison
    - BERT
    - SciBERT
    - BioBERT
    - PubMedBERT
- BioBERT intermediate binary model
- BioBERT final model
    - Evaluation against benchmarks
    - Comparison to published models

### 4.2 Overall Results – non-LLM Models

After preprocessing train/test data, the logistic regression, random forest, and gradient boosting models were built and implemented as described in Chapter 3. The results are shown in Table 4-1 and Figure 4-1, below.

**Table 4-1. Performance Metric Means Across All Train/Test Sets**

| Models | ROC-AUC | Bal'd Acc | Precision | Recall | F1 | PR-AUC |
|---|---|---|---|---|---|---|
| Logistic Regression | **0.639** | **0.598** | **0.877** | 0.496 | 0.623 | 0.875 |
| Random Forest | 0.636 | 0.595 | 0.866 | 0.570 | 0.681 | 0.873 |
| Gradient Boosting | **0.639** | 0.597 | 0.864 | **0.596** | **0.698** | **0.876** |

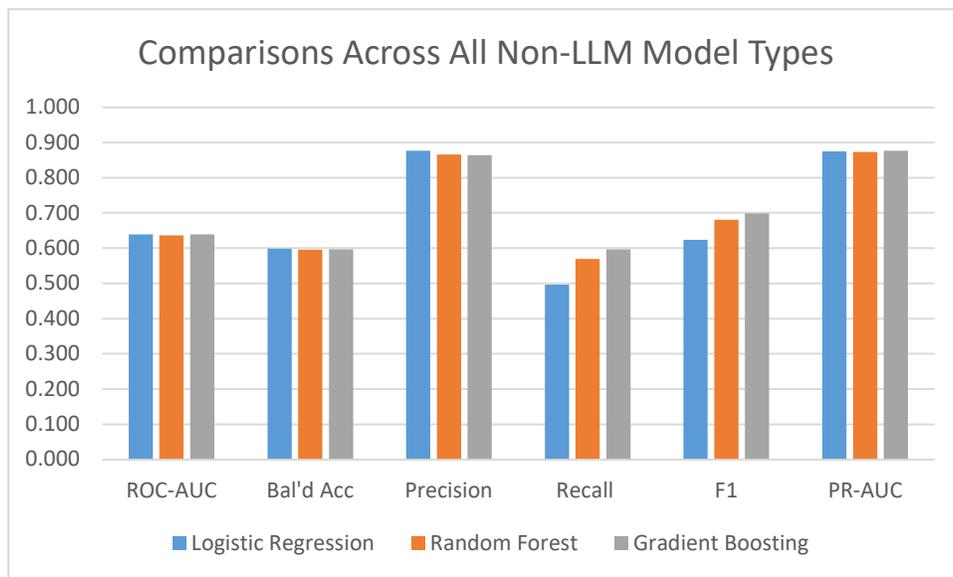

**Figure 4-1. Comparisons of Mean Performance for each non-LLM Model Type**

Each of the three models achieved scores above 0.50 in balanced accuracy, ROC-AUC, and PR-AUC, indicating that they performed better than would be expected from random guessing. This also appears to satisfy Hypothesis 1 (Chapter 1). The logistic regression model performed marginally better in balanced accuracy and precision, while the gradient boosting model demonstrated significantly better performance in recall (0.596, vs. 0.496 for logistic regression) and F1 (0.698, vs. 0.623 for logistic regression), and marginally better in PR-AUC. Whereas both ROC-AUC (for which both gradient boosting and logistic regression scored a 0.639) is a more common metric used in predictive model evaluation, PR-AUC is better at addressing the issue of imbalanced datasets. While these results do not indicate overwhelming superiority for the gradient boosting model, it does appear as though it was the better of the three. Because of this, the gradient boosting model was selected for further interrogation using feature extraction.

**Gradient Boosting Feature Extraction**

The purpose of feature extraction is to define the input parameters or tokens (features) which are most impactful for model prediction, and quantify that impact through a SHAP value, as explained in Chapter 2.

For Phase 1, all SHAP values were negative, indicating that each of the 20 most prominent features were associated with a reduction in the predicted pTRS for the program. This indicates that penalizing trial descriptions for key elements may be a more effective means by which to make good predictions, as opposed to rewarding them for more auspicious elements.

Among the most impactful terms were 'survival,' perhaps because trials in the test set with descriptions that included survival as a clinical endpoint were less likely to be successful. "Cancer" was another inauspicious term, as well as "ctcae." CTCAE is a grading system used to evaluate adverse events in cancer interventions, and thus bears tight correlation with the "cancer" token (Trotti, et al., 2003). A diagram showing the top 20 impactful tokens and their SHAP values can be seen in Figure 4-2.

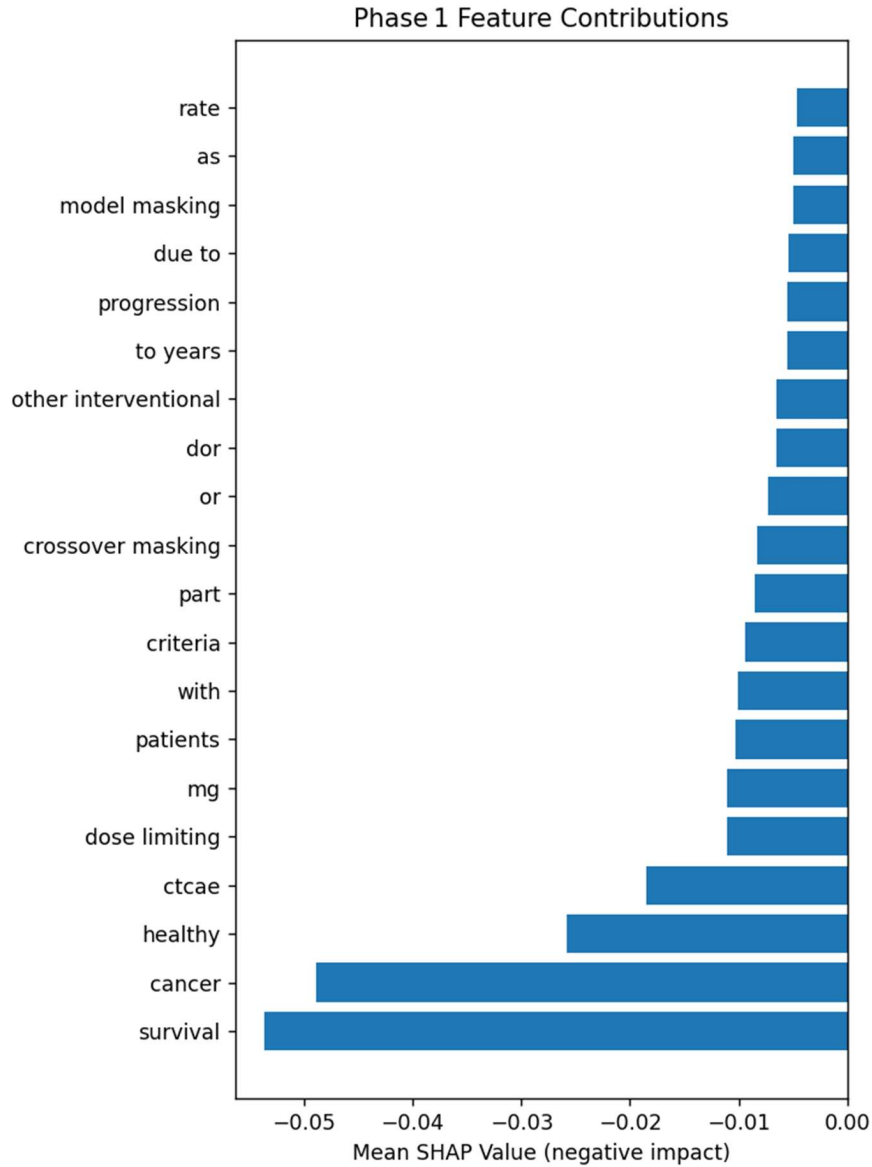

**Figure 4-2. SHAP Values for Phase 1**

Figure 4-3, below, shows the feature contributions for the 20 most impactful tokens in Phase 2. In this case, one term, "terminated," was found to have a positive correlation with a higher success prediction. "Free survival," the most inauspicious token for Phase 2, is most likely a fragment of the parent term "progression-free survival," a common clinical endpoint which refers to the length of time that a patient lives with a disease but it does not progress (Gyawali, et al., 2022). This seems to indicate that trials for which this endpoint is important may be correlated with lower probabilities of success (at least, in this model).

"Allocation intervention" appears to be referring to how trial participants are allocated into different interventional groupings, while "mechanical ventilation" could be a significant negative factor because of the number of COVID-19 treatment candidates that were in development, many unsuccessfully, from 2020 to 2024.

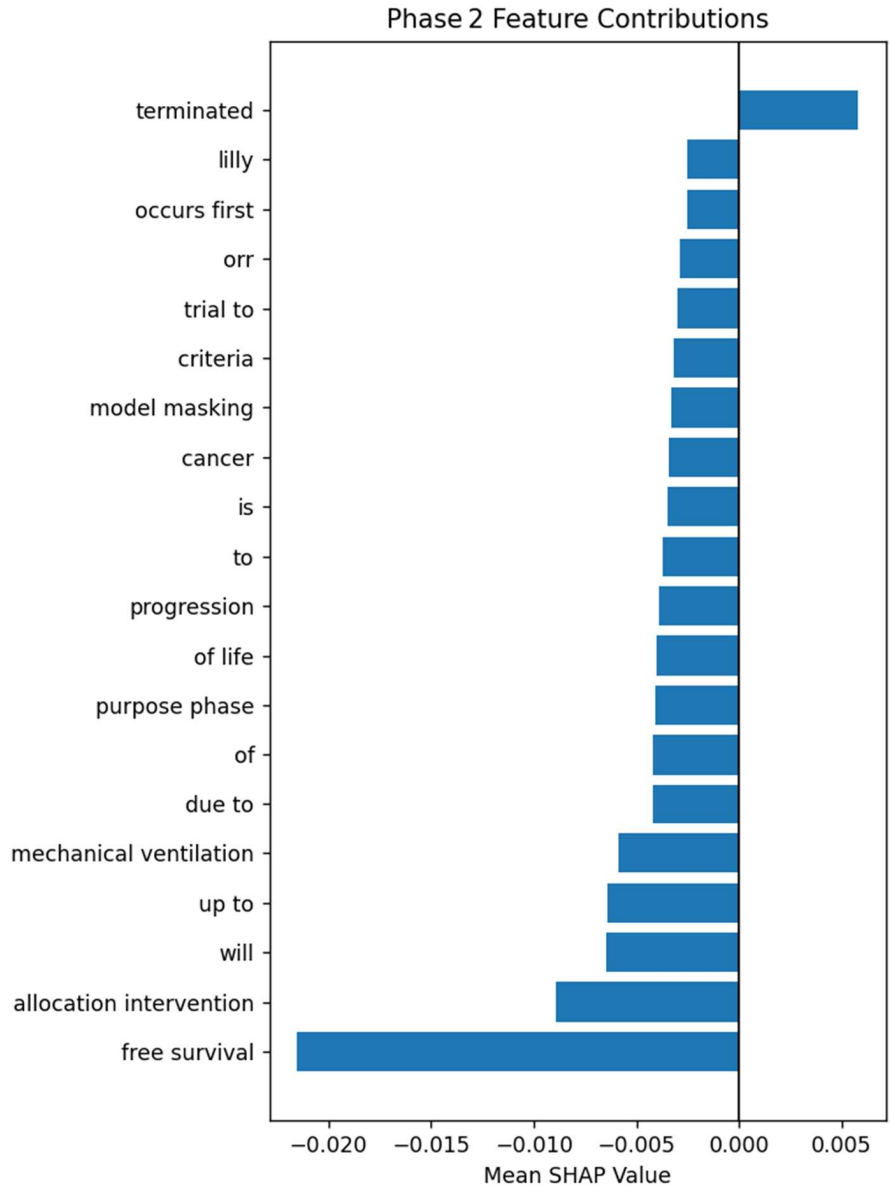

**Figure 4-3. SHAP Values for Phase 2**

The results of the feature extraction analysis for Phase 3 are shown in Figure 4-4. Here, we see that six of the 20 most prominent tokens carry positive values, indicating that they are correlated with a higher predicted pTRS. Several of these, such as "pfizer investigational," "vaccine," and "vaccination," may be related to COVID-19 vaccines, several of which were successfully approved and launched.

Interestingly, "progression" is listed as a token correlated with lower pTRS predictions, potentially because of its use in the phrase "progression free survival," as seen in the Phase 2 feature extraction. This would indicate that this term is potentially a powerful negative predictor across multiple phases. More of the tokens in Phase 3 appear to be associated with locations in which studies were conducted, perhaps due to significant historical differences in study outcomes in these locations.

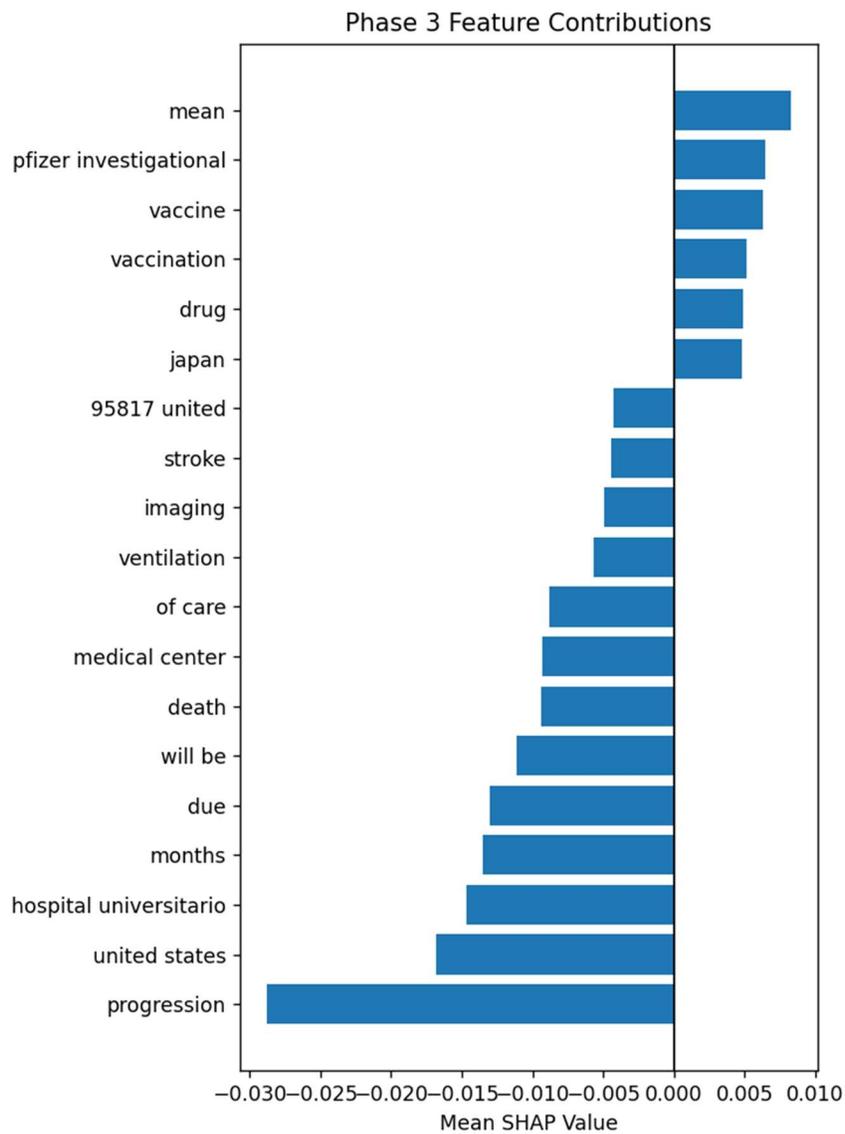

**Figure 4-4. SHAP Values for Phase 3**

**Comparison of All Indication vs. Neuroscience Only Train/Test Sets**

One goal for this praxis was to evaluate the validity of Hypothesis 3, which pertains to which training set is better to use, one containing trial descriptions for all indications, or one containing only those for neuroscience indications. This is difficult to surmise intuitively, since the 'all indication' dataset contains more trials to train on, but the neuroscience-only dataset contains training data that may be more concentrated in the most pertinent features for pTRS prediction.

As the data in Table 4-2 and Figure 4-5 demonstrate, when averaging across all phases and models, the use of all trial indications to train the model results in better performance metrics across the board, and particularly for recall and F1. This appears to indicate that the additional trials available result in a more robust overall model, even if the training set is no longer specific to the neuroscience therapeutic area.

Table 4-2. Means of non-LLM Model Metrics by Training/Test Sets

| Training Set | ROC-AUC | Bal'd Acc | Precision | Recall | F1 | PR-AUC |
|---|---|---|---|---|---|---|
| All Indications Mean | **0.641** | **0.600** | **0.875** | **0.592** | **0.698** | **0.882** |
| Neuroscience Only Mean | 0.635 | 0.593 | 0.863 | 0.516 | 0.637 | 0.868 |

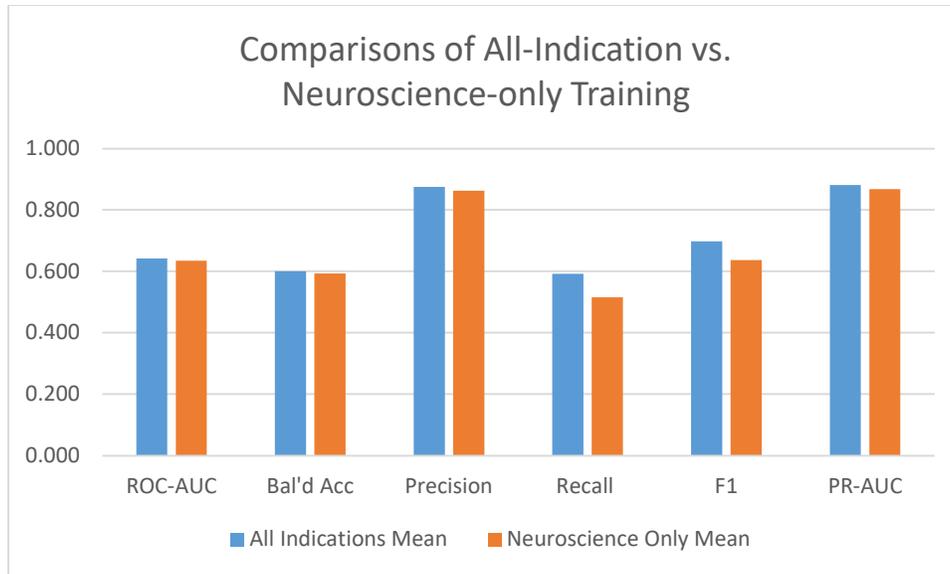

**Figure 4-5. Performance Metrics for All Indications and Neuroscience-only Training Sets**

**Comparisons by Phase**

Another important question is how applicable the predictive model is in different phases. Each phase has different criteria for success, and a model that is predictive in only some of those phases will have a significant weakness. As shown in Table 4-3 and Figure 4-6, the model fared slightly better in predicting the success of Phase 1 trials than Phase 3 trials, but both of those were far better than the predictions for Phase 2 trials. In fact, it appears that Phase 2 predictive validity is enough of an outlier to be dragging down the means across the board.

**Table 4-3. Non-LLM Performance Metrics by Phase**

| Phase | ROC-AUC | Bal'd Acc | Precision | Recall | F1 | PR-AUC |
|---|---|---|---|---|---|---|
| Phase 1 | **0.662** | 0.613 | **0.886** | **0.668** | **0.760** | **0.899** |
| Phase 2 | 0.595 | 0.564 | 0.846 | 0.384 | 0.525 | 0.838 |
| Phase 3 | 0.657 | **0.614** | 0.874 | 0.610 | 0.717 | 0.886 |

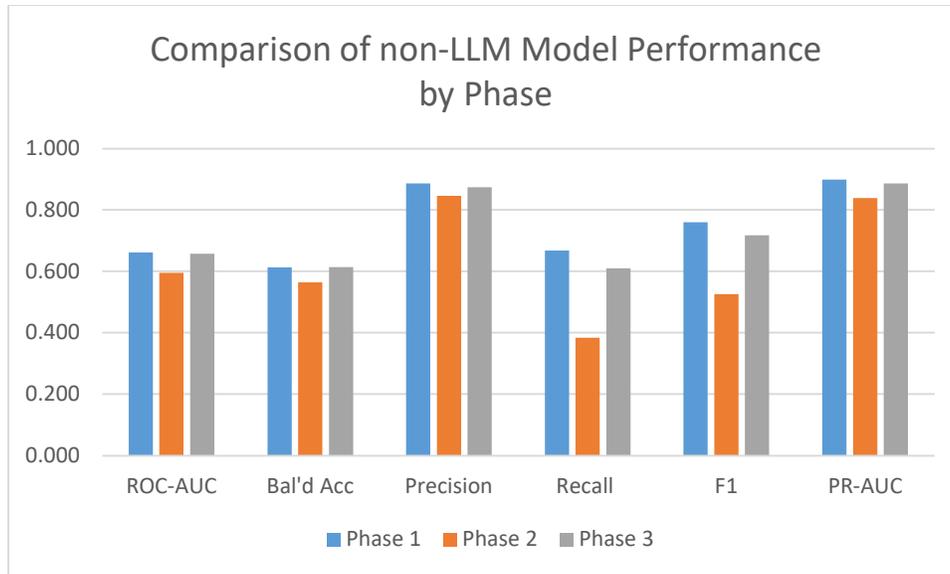

**Figure 4-6. Non-LLM Performance Metrics by Phase**

Additional data and figures from the non-LLM models can be found in Appendix A

## 4.3 LLM Transformer Comparison

The next step in this work was to determine whether LLM/transformer-based approaches could surpass the performance of the non-LLM models. The first step was to identify the most suitable LLM/transformer model to use. Table 4-4 and Figure 4-7 show the performance metrics for an initial evaluation in which four related but different LLMs were evaluated: BERT, BioBERT, SciBERT, and PubMedBERT.

**Table 4-4. LLM Performance Metrics by Transformer**

|            | ROC-AUC | Bal'd Acc | Precision | Recall | F1    | PR-AUC |
|------------|---------|-----------|-----------|--------|-------|--------|
| BERT       | 0.595   | 0.537     | 0.833     | 0.860  | 0.846 | 0.871  |
| BioBERT    | 0.604   | 0.530     | 0.835     | **0.908** | **0.87** | 0.87   |
| SciBERT    | **0.637** | **0.554** | **0.839** | 0.868  | 0.853 | **0.9** |
| PubMedBERT | 0.628   | 0.537     | 0.833     | 0.868  | 0.85  | 0.885  |

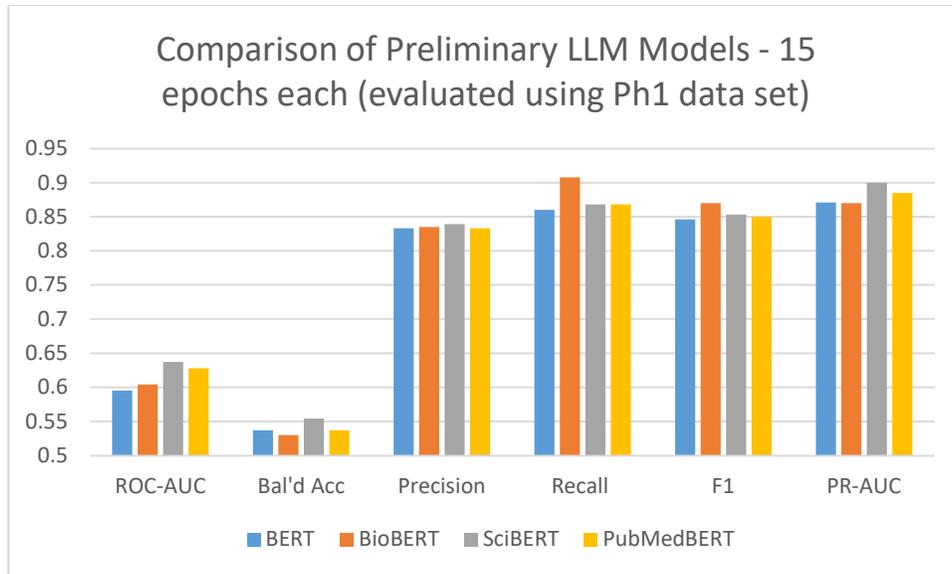

**Figure 4-7. LLM Performance Metrics by Transformer**

As shown, BioBERT and SciBERT each performed well in these metrics. Ultimately, because BioBERT was trained specifically on biomedical corpora (narrower in scope and better aligned with clinical trials than SciBERT), BioBERT was selected as the LLM to move forward with. All models in the succeeding sections were built using BioBERT.

## 4.4 BioBERT Initial Model Performance

Initially, the BioBERT models were built by training and testing on the entire dataset, for all indications, for each phase. The models were then trained and tested again using the smaller, neuroscience-only datasets. As shown in Table 4-5 and Figure 4-8, the all-phase means of 'all indication' train/test models outperformed the 'neuroscience only' train test models in every metric except for recall (which was quite close).

**Table 4-5. BioBERT Initial Model Performance**

| Model | ROC-AUC | Bal'd Acc | Precision | Recall | F1 | PR-AUC |
|---|---|---|---|---|---|---|
| **All Indications Phase 1** | **0.743** | 0.559 | **0.864** | 0.905 | **0.884** | **0.94** |
| **All Indications Phase 2** | 0.671 | 0.554 | 0.811 | 0.847 | 0.829 | 0.894 |
| **All Indications Phase 3** | 0.707 | **0.583** | 0.855 | 0.872 | 0.864 | 0.923 |
| **Neuroscience Only Phase 1** | 0.604 | 0.531 | 0.835 | **0.908** | 0.870 | 0.870 |
| **Neuroscience Only Phase 2** | 0.592 | 0.550 | 0.790 | 0.743 | 0.766 | 0.819 |
| **Neuroscience Only Phase 3** | 0.639 | 0.567 | 0.824 | 0.822 | 0.823 | 0.869 |
| **Overall Mean** | 0.659 | 0.557 | 0.830 | 0.850 | 0.839 | 0.886 |

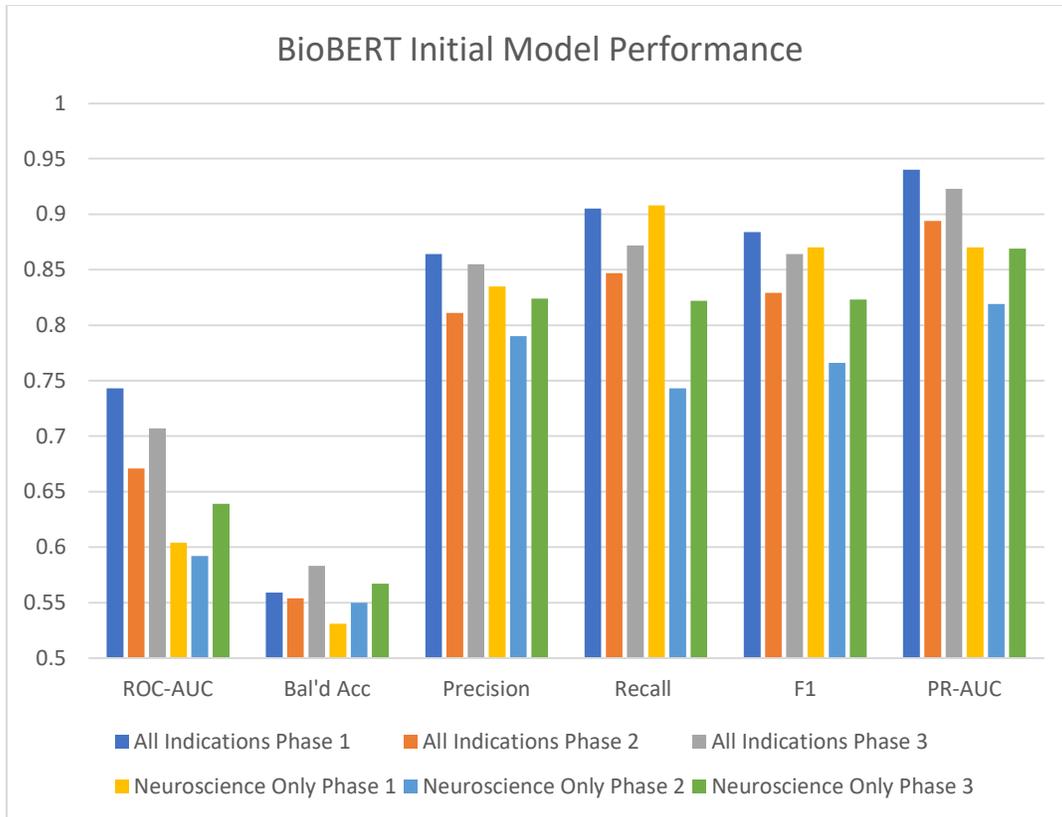

**Figure 4-8. BioBERT Initial Model Performance**

The next step was to compare the all indication and neuroscience-only model performance means to models which were trained on all clinical trials but tested only on neuroscience. This comparison, shown in Table 4-6 and Figure 4-9, shows that the all-phase mean of the "Train All Test Neuro" combination was superior to the other model means in most metrics, and very close in precision and PR-AUC.

**Table 4-6. BioBERT Model Performance Metrics by Training/Test Set**

| Train/Test Set | ROC-AUC | Bal'd Acc | Precision | Recall | F1 | PR-AUC |
|---|---|---|---|---|---|---|
| **All Indications** | 0.707 | 0.565 | **0.843** | 0.875 | 0.859 | **0.919** |
| **Neuroscience Only** | 0.612 | 0.549 | 0.816 | 0.824 | 0.820 | 0.853 |
| **Train All Test Neuro** | **0.740** | **0.573** | 0.832 | **0.900** | **0.864** | 0.904 |

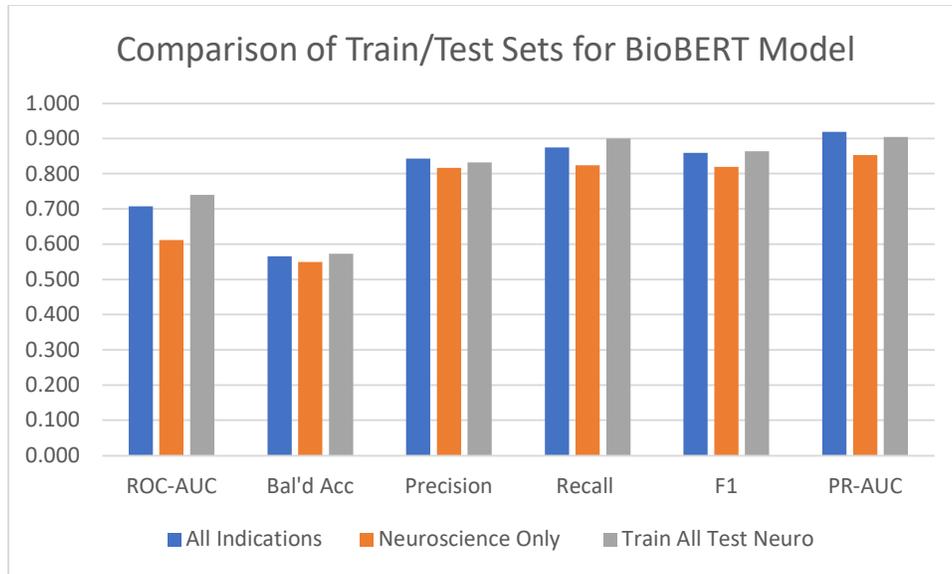

**Figure 4-9. BioBERT Binary Model Performance Metrics by Training/Test Set**

## 4.5 BioBERT Final Model Performance

All final models were evaluated using the BioBERT LLM, with models trained on all clinical trials in the dataset and tested on only neuroscience trials. As described in Section 3.6 of the Methodology, the final models used continuous probability labels from the CTO Dataset, rather than rounding to a binary label. The Phase 1-3 breakdown is shown below in Table 4-7 and Figure 4-10. Phase 1 shows significantly superior performance metrics to phases 2 and 3 in ROC-AUC, balanced accuracy, and PR-AUC, similar to the trend that was seen in the non-LLM models.

Figure 4-11 shows the ROC-AUC curves which correspond to the scores in Table 4-7. Here, the dashed lines indicate model performance akin to guessing, and greater area under the black curve corresponds to a model that is both sensitive enough to capture true positives and specific enough to avoid false positives.

Table 4-7. BioBERT Final Model Performance Metrics

| Model | ROC-AUC | Bal'd Acc | Precision | Recall | F1 | PR-AUC |
|---|---|---|---|---|---|---|
| **Train All Test Neuro Phase 1** | **0.843** | **0.601** | **0.840** | 0.900 | 0.869 | **0.949** |
| **Train All Test Neuro Phase 2** | 0.733 | 0.597 | 0.815 | 0.829 | 0.822 | 0.884 |
| **Train All Test Neuro Phase 3** | 0.644 | 0.520 | **0.840** | **0.970** | **0.900** | 0.879 |
| **Mean Across Phases** | 0.740 | 0.573 | 0.832 | 0.900 | 0.864 | 0.904 |

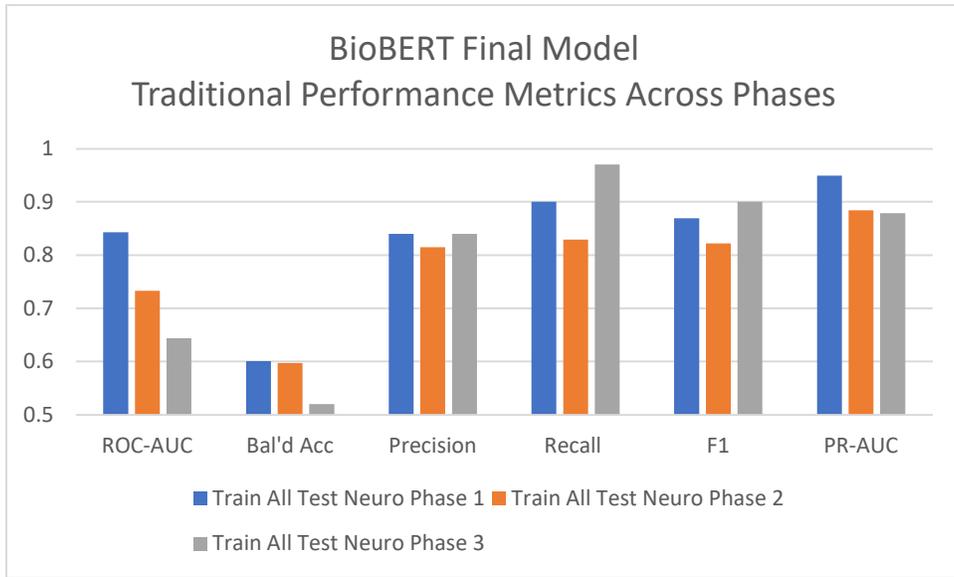

Figure 4-10. BioBERT Final Model Performance Metrics

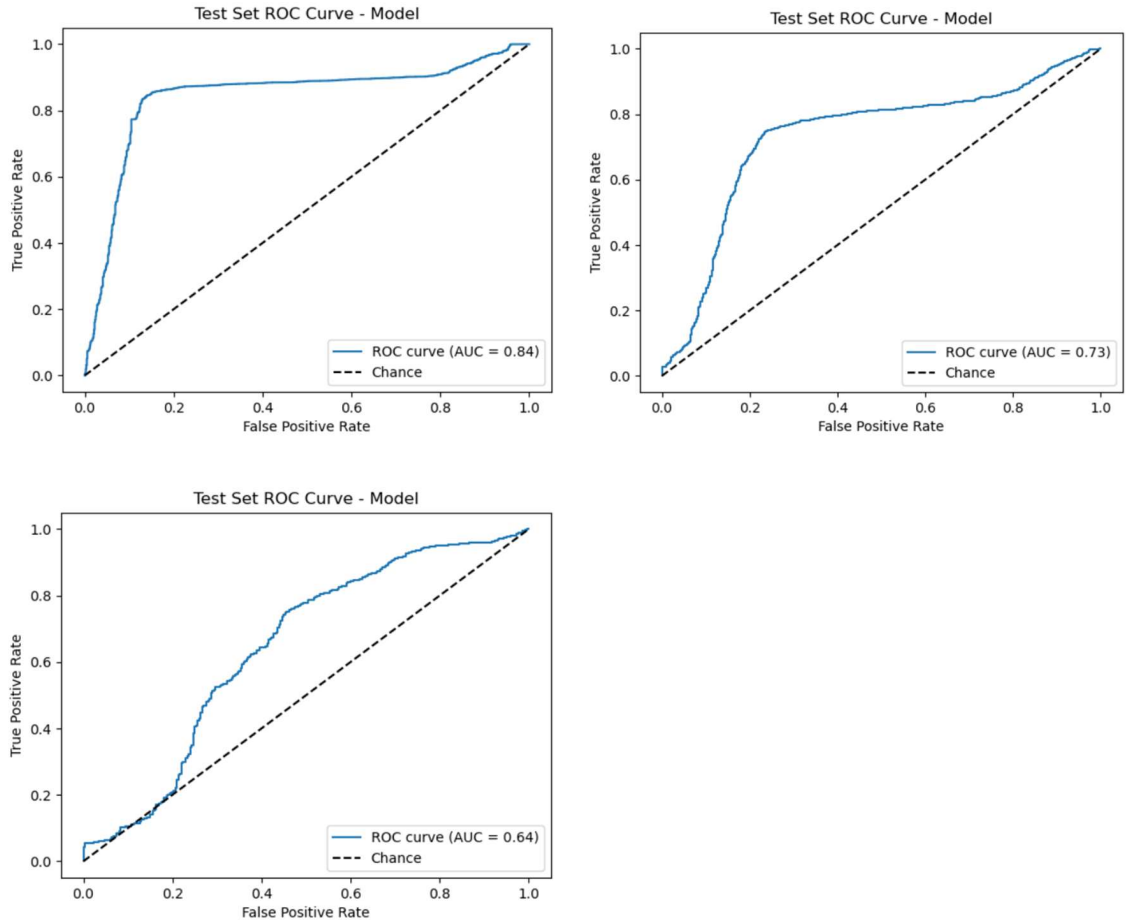

**Figure 4-11. ROC Curves for BioBERT Model Predictions**
*Top Left: Phase 1. Top Right: Phase 2. Bottom Left: Phase 3.*

Additional Metrics of interest for comparing the distance between predicted (model) and actual (label) values are log loss and Brier score, described in greater depth in Chapter 2. For these metrics, lower values represent superior performance. As shown in Table 4-8 and Figure 4-12, Phase 1 and 3 performance was quite similar, while phase 2 predictions appeared to fare worse by these metrics.

Table 4-8. BioBERT Final Model Additional Metrics

| Model | Log Loss | Brier Score |
|---|---|---|
| **Train All Test Neuro Phase 1** | 0.637 | 0.163 |
| **Train All Test Neuro Phase 2** | 0.956 | 0.238 |
| **Train All Test Neuro Phase 3** | **0.634** | **0.155** |

*Note*: For Log Loss and Brier Score, lower values represent superior performance.

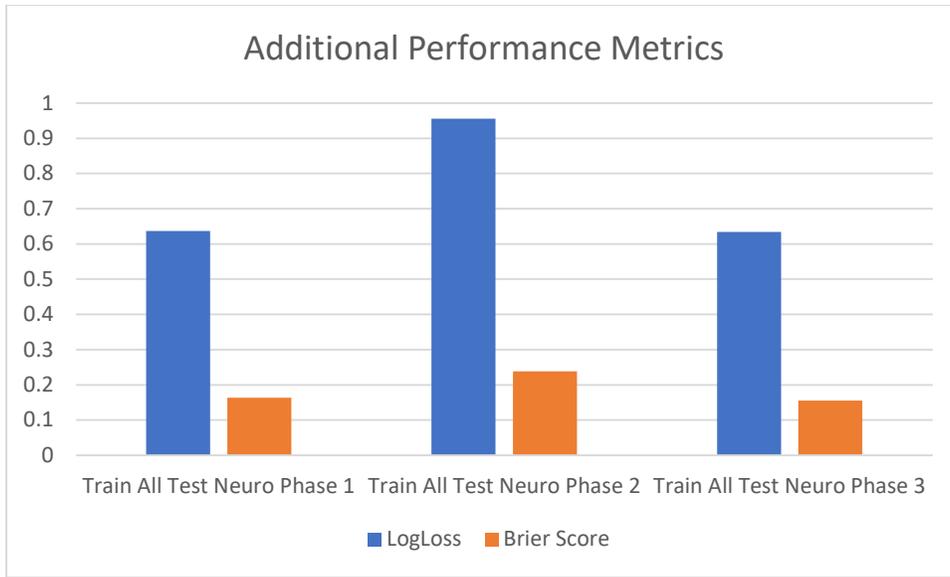

Figure 4-12. BioBERT Final Model Additional Metrics

## 4.6 BioBERT Final Model Performance vs. Benchmarks

Finally, the final models for Phases 1-3 were evaluated with respect to the industry and dataset benchmarks. As a reminder, the industry benchmark is the one provided by Citeline, as described in Sections 3.2 and 3.6, and provides industry average trial outcome pTRS values. The dataset benchmark, also described in Chapter 3, is the actual mean outcome across all trials within the dataset that the model was trained on (this disadvantages the model, since this mean includes outcomes of trials that the model was tested against, not just that it was trained on). Superiority to the industry benchmark indicates an improvement over common industry practice, whereas superiority to the dataset benchmark would represent excellent or aspirational performance. The evaluation was conducted to see if the model's individual trial predictions would be superior to simply using one of the benchmarks instead for all trials.

As shown in Table 4-9 and Figure 4-13, the log loss and Brier score for the final model were both lower (superior) to the dataset benchmark in all three phases. Additionally, the proportion of trials in which the model made a superior prediction (one that was closer to the outcome label) was measured, and compared to the benchmarks as shown in Table 4-9 and Figure 4-14. The industry benchmark was surpassed by the model for all phases, and only for Phase 3 did the model fail to surpass the aspirational bar of the dataset benchmark, although Phase 3 all-phase mean performance for the model was still superior to the dataset benchmark.

**Table 4-9. BioBERT Performance Metrics vs. Benchmarks**

| Model | Log Loss | Brier Score | % of time model beats benchmark |
|---|---|---|---|
| **Model Phase 1** | 0.637 | **0.163** | n/a |
| **Industry Benchmark Ph 1** | 0.716 | 0.261 | **76%** |
| **Dataset Benchmark Ph 1** | **0.565** | 0.187 | **79%** |
| **Model Phase 2** | 0.956 | 0.238 | n/a |
| **Industry Benchmark Ph 2** | 1.034 | 0.408 | **71%** |
| **Dataset Benchmark Ph 2** | **0.629** | 0.218 | 62% |
| **Model Phase 3** | 0.634 | **0.155** | n/a |
| **Industry Benchmark Ph 3** | 0.687 | 0.247 | **64%** |
| **Dataset Benchmark Ph 3** | **0.53** | 0.17 | 48% |
| **Mean Using Model** | 0.742 | **0.185** | n/a |
| **Mean Using Industry Benchmark** | 0.812 | 0.305 | **70%** |

| Mean Using Dataset Benchmark | 0.575 | 0.192 | 63% |

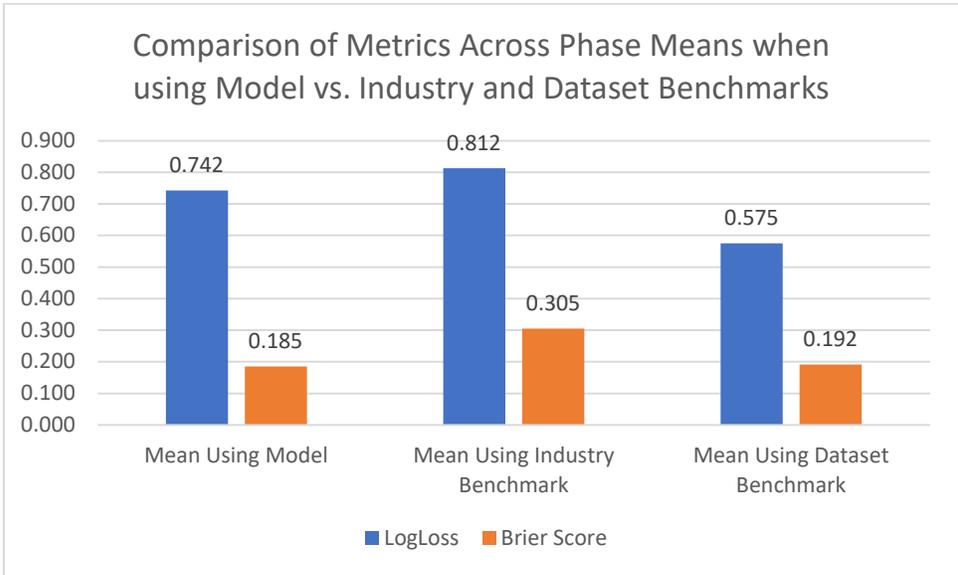

Figure 4-13. BioBERT Performance Metrics vs. Benchmarks

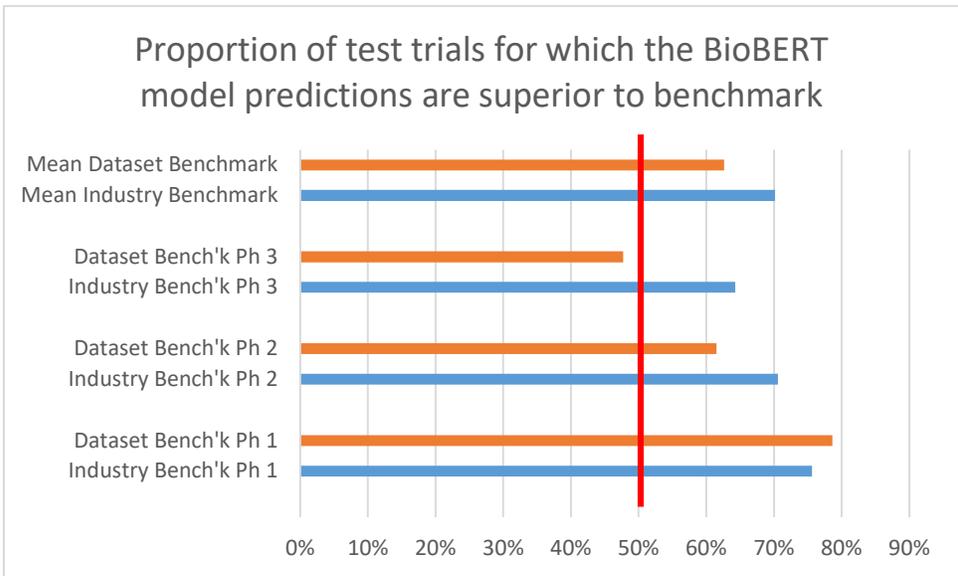

Figure 4-14. Proportion of Trials Where Model Outperformed Benchmarks

4.7 Comparison of BioBERT Final Model to Published Models

In order to understand the performance of the final BioBERT LLM model relative to those described in published work with similar aims, Table 4-10 was assembled. ROC-AUC was the most consistently available metric that would allow for a comparison of both the sensitivity and specificity of the models. The model described here outperformed most of the other models reporting Phase 1 scores, and across all phases, the mean ROC-AUC score for this model comes in tied for #4 out of 7, despite being the only one to use no drug characteristic information other than what is contained in ClinicalTrials.gov.

**Table 4-10. Comparison of Model ROC-AUC Relative to Published Methods**

|  | ROC-AUC Score | | | |
| --- | --- | --- | --- | --- |
|  | Phase 1 | Phase 2 | Phase 3 | Mean |
| **This Work** | 0.74 | 0.67 | 0.71 | 0.71 |
| Lo, et al. (2017) | n/a | 0.78 | **0.81** | 0.8 |
| Gao, et al (2024) | **0.77** | 0.74 | 0.75 | 0.75 |
| Feijoo, et al. (2020) | n/a | 0.74 | 0.67 | 0.71 |
| Fu, et al. (2020) | 0.58 | 0.65 | 0.72 | 0.65 |
| Aliper, et al. (2023) | n/a | **0.88** | n/a | **0.88** |
| Zheng, et al. (2024) | 0.65 | 0.65 | 0.74 | 0.68 |

**4.8 Summary**

The results of this praxis were able to satisfactorily answer all three of the research questions from Section 1.6:

**RQ1:** Can a non-LLM NLP model be built that can generate >5% better-than-random pTRS predictions for neuroscience indications?

Yes. Accuracy relative to random predictions is best measured by balanced accuracy. As demonstrated in Section 4.2, balanced accuracy metrics for all three non-LLM models were above 59% (50% would correspond to random guessing).

**RQ2:** Does the use of an LLM-enhanced NLP model lead to a >5% improvement in pTRS prediction over non-LLM for neuroscience indications?

Yes. In terms of quality of prediction, as measured by ROC-AUC, the all-phase mean ROC-AUC of 0.74 for the BioBERT model is a 15.8% improvement over the gradient boosting ROC-AUC of 0.639.

**RQ3**: Can a model trained solely on ClinicalTrials.gov data and outcome labels surpass the predictive performance of industry benchmarks by >5% for neuroscience indications?

Yes. As shown in Section 4.6, the final BioBERT model made better predictions of trial outcome than industry benchmarks for 70% of trials in the test set, with a mean Brier score 39.9% lower (better) than when benchmarks were used for the pTRS predictions.

**RQ4:** Does use of training data from all indications (not just neuroscience) improve neuroscience pTRS prediction >5%?

Yes, training the BioBERT model on trials from all indications resulted in a mean ROC-AUC score of 0.740, a 20.9% improvement over the 0.612 ROC-AUC seen when training only on trials for neuroscience indications (See Table 4-6).

## Chapter 5—Discussion and Conclusions

**5.1 Discussion**

Pharmaceutical R&D involves the meticulous curation of programs within companies' development pipelines. Extensive analysis is completed in order to ensure that the most promising drug candidates are provided with the financial resources to complete costly clinical trials. This analysis includes the calculation of expected net present value (eNPV), and other metrics to define the risk adjusted value of programs. These risk adjusted values ultimately hinge on one extraordinarily important estimate: Probability of Success, or pTRS. Because of trial expense (hundreds of millions of dollars) and enormity of potential returns (billions of dollars annually), even small changes to a program's pTRS estimates may result in large shifts in the expected value of these programs, and thus alter decisions on whether to proceed with or terminate a drug candidate's development. Because of this importance, drug companies spend significant time (developing estimates), and money (investing in proprietary datasets) to improve their predictions to allow for better decisions to be made.

In order to build a model to generate predictions superior to industry benchmarks, this praxis leveraged publicly available data from ClinicalTrials.gov. The data was downloaded, sorted, and cleaned to prepare it for analysis. Any categories of information (such as patient number) that could have leaked data about trial status or outcome during training was removed. Clinical Trial Outcomes Database (CTOD) data from Gao, et al. (2024) was then also downloaded, and NCT (clinical trial ID) numbers were used to link clinical trial outcome labels with their corresponding trials in the clinical trial data CSVs.

A subset of the comprehensive "all indication" train/test set was created containing only neuroscience trials. Industry benchmarks, providing the mean probabilities of trial success for different clinical phases, were downloaded from Citeline, and dataset benchmark means were calculated based on the labels within the train/test sets used.

Non-LLM models were built in order to evaluate random forest, gradient boosting, and logistic regression approaches. These were evaluated in terms of ROC-AUC, PR-AUC, balanced accuracy, recall, precision, and F1. Each was able to make predictions far superior to random guessing, with the gradient boosting model being the best across many metrics.

An LLM-based approach was then implemented, centering around the BERT (Bidirectional Encoder Representations from Transformers) family of transformer models. Several BERT types were evaluated, with little clear difference among them, and BioBERT was chosen for use in building a refined LLM-based model. The final model was able to produce a better prediction than the industry benchmark in 70% of trials, with a mean Brier score 40% lower (better) than the industry benchmark.

This praxis sought to determine whether a useful model could be built to make successful predictions of clinical trial outcomes based solely upon their descriptions in ClinicalTrials.gov. The criteria for this evaluation (Section 4.8) included greater than 5% improvement over random guessing and improvement over industry benchmarks. The final model described in this praxis was able to successfully surpass these thresholds.

As illustrated in Section 4.7, the model described in this praxis was outperformed by those built by some research teams However, those teams leveraged data such as molecular profiles from manually curated databases of bioactive molecules, and/or drug compound attributes from *Pharmaprojrcts* profiles database. Both of these databases were used in order to train models on information about the drug and/or target molecules themselves to enhance predictions. Likewise, Feijoo, et al., used proprietary Biomedtracker data to supply information about disease area and drug class for each trial in their training set. Fu, et al., used molecular and pharmacokinetic information about the drugs used in trials from MoleculeNet for their training sets, and Zheng, et al, used SMILES, string-format information about molecular structures. Since the aim of this praxis was to build a model that did not use databases other than ClinicalTrials.gov and the outcome labels, the model described herein was surprisingly competitive against these other models that leveraged additional molecular data.

*Limitations*

A key limitation of this work is that the clinical trial outcome dataset (CTOD) labels are not, themselves, manually curated. Though they leverage large amounts of data that appears after clinical trial completion, inaccuracies can exist in that dataset which could cost overall model accuracy. In this way, there exists a potential tradeoff between using a large set of outcome labels (like CTOD) that may have inaccuracies, or instead training a model with human-curated train/test datasets where the confidence in the label accuracy is greater, as other authors have done.

Clinical trial descriptions from ClinicalTrials.gov that were missing critical elements needed for processing were omitted. If those omissions fit some pattern (e.g., if poorly funded trials positively correlate with incomplete trial data) this could potentially bias the model.

The models described in this praxis were also trained on clinical trial data in a very specific format – the fields associated with a clinical trial description published within the database. A company would therefore need to have a description set that aligns very closely with this in order for the model to be able to make appropriate predictions, and that is not possible until a drug candidate's clinical trial has been designed (see Section 3.7). If this information is not available from ClinicalTrials.gov or a similar database, but a trial has been designed, a team could choose to request that the necessary data fields (e.g., "Summary," "Conditions," "Interventions") be drafted by their clinical development colleagues in order to facilitate pTRS estimation. However, it should be noted that for the model's predictions to be valid and applicable, the individuals providing content for these data fields *must have no knowledge* of trial properties and descriptors that correlate with higher or lower success predictions (e.g., the SHAP values from Section 4.2). This could very easily bias the model inputs and invalidate the predicted pTRS.

Along a similar line, it may be of interest to clinical development teams to understand the specific tokens that were correlated with higher outcome predictions. This might lead them to focus on clinical approaches that appeared to be more auspicious. But they should understand that the interactions between tokens and outcomes may be very complex, and correlation does not necessarily imply causation. Currently, it would be difficult to ascertain SHAP-like values from the BioBERT model, since it is based on a complex neural network transformer, which makes it more difficult to reverse engineer potentially fortuitous trial properties.

Potential users of this model should also be aware that while the clinical trial data it was trained on contains significant information about the trial, its endpoints, location, target patient population, etc., there are simply many variables that could result in the termination of a trial or drug development program that are not ever going to be known at the time the clinical trial description is written. Some safety concerns cannot be identified in any way until a drug has been administered to a sufficiently large human patient population, and drug efficacy is dependent on a myriad of molecular and systemic interactions that the trial descriptions simply cannot shed light on. For these reasons, no predictive pTRS model will ever be able to approach 100% accuracy.

**5.2 Conclusions**

Pharmaceutical innovation is inherently limited by risk, and the enormous expense of failed clinical trials. Because of this, it is imperative that drug developers are able to scope probability of technical and regulatory success as accurately as possible. This work demonstrates that non-LLM NLP models can provide better than random predictions of pTRS based only on ClinicalTrials.gov data. This allows for useful models to be used by companies and organizations who may not have access to expensive and specialized technical data or GPU hardware to train pTRS prediction models. An LLM-based model was then built that generated even better predictions, able to significantly surpass the predictive validity of industry benchmarks for neuroscience trials. Since industry benchmarks are often used by companies to help them understand technical risk and make go/no-go decisions, the predictive improvements demonstrated in this work can enable better asset allocation decisions.

**5.3 Contributions to Body of Knowledge**

1. This praxis describes a novel use of the Clinical Trial Outcome Dataset to aid pTRS prediction, demonstrating the utility of highly comprehensive outcome label sets.

2. This praxis utilizes only ClinicalTrials.gov for training data, demonstrating that significant insight can be gained without deep background on drug molecular structure. This underlying training data framework can then be scaled with whatever additional data a team does have access to in order to further optimize predictions.

3. This praxis demonstrates, using neuroscience as an example indication, that training a single-indication predictive pTRS model with clinical trials from all indications may produce better results than a narrower training approach using only the target indication.

4. This praxis showed that pTRS predictions for neuroscience, a therapeutic area with unique drug delivery and toxicological challenges, could be improved immensely with a model trained on open-source data.

**5.4 Recommendations for Future Research**

The following activities could enhance the predictive power and utility of the final model:

1. *Evaluate additional LLM transformer models*

    Other LLMs, such as BioGPT, or GatorTron from the University of Florida, could exceed the predictive performance of the BioBERT model.

2. ***Test the model against one of the smaller, human-curated trial outcome label sets***

   This could provide information on the impact of any potential inaccuracies in the labels from the CTOD dataset.

3. ***Identify key tokens from the BioBERT or other LLM model found to be correlated with higher or lower pTRS estimates***

   While this was done for the gradient boosting model, this is more challenging for a neural network based LLM. However, these insights could be particularly helpful, aiding development teams in identifying auspicious trial features.

4. ***Evaluate the model across all individual therapeutic areas***

   Since neuroscience was the only indication evaluated, additional insights could be gleaned from other therapeutic areas such as cardiovascular or oncology drugs. This would also help to evaluate whether the finding that predictions are improved by using all indications to train the model are generalizable to other therapeutic areas.

5. ***Evaluate a version of the model that uses all phases, with phase number as a predictive feature***

   Currently, there is a separate model for all three phases. By building one large comprehensive model and then having Phase number be a selectable feature, it would be possible to leverage the information from trials in other phases, which could potentially improve predictions.

6. ***Evaluate how the model performs on data from different time periods, and identify trends***

It may be that the model performs better on trials from a certain period, or performs better when not trained on much older clinical trial data. Adjusting time parameters for train/test sets should be done in a way that maintains the sequential fidelity of training and testing, so that the model is not trained on any trials that occurred after any of the test set trials.

7. ***Identify additional open-source data that could be easily integrated into the train/test data***

   While no predictive pTRS model will be able to make perfect predictions, adding features to the current model, such as pharmacological data, could enhance its predictive utility.

8. ***Conduct a post-hoc analysis on the best and worst predictions***

   Across the test set, it would be possible to identify the best predictions (those predictions for which the model's success probability prediction values were closest to their corresponding 0/1 binary success/fail outcome) and the worst. Exploratory data analysis could then be conducted to identify patterns that could lead to model improvements.

9. ***Enable data pipelines that allow for continuous updating of model training***

   The biotechnology regulatory landscape can evolve relatively quickly, so an agile model that can be re-trained at a regular cadence would enhance pTRS prediction fidelity. Gao, et al., have included tools in their GitHub to facilitate user-updating of clinical trial outcome estimates for emerging trials, meaning that labels could be available for regular re-training of the model over time.

Continual, incremental improvements to this model may pave the way for a predictive tool that allows companies to make better allocation of capital resources, so that they can improve the arsenal of therapeutic tools available to help patients.

# References

Aggarwal, A., Xu, Z., Feyisetan, O., & Teissier, N. (2020). On Log-Loss Scores and (No) Privacy. *Proceedings of the Second Workshop on Privacy in NLP*, 1–6. https://doi.org/10.18653/v1/2020.privatenlp-1.1

Aliper, A., Kudrin, R., Polykovskiy, D., Kamya, P., Tutubalina, E., Chen, S., Ren, F., & Zhavoronkov, A. (2023). Prediction of Clinical Trials Outcomes Based on Target Choice and Clinical Trial Design with Multi-Modal Artificial Intelligence. *Clinical Pharmacology & Therapeutics*, *114*(5), 972–980. https://doi.org/10.1002/cpt.3008

Andersen, J. (2012). *Probability Elicitation and Calibration in a Research & Development Portfolio: A 13-Year Case Study.* [PowerPoint Slides]. Eli Lilly and Company. https://slideplayer.com/slide/8394454/

Austin, D. & Hayford, T. (2021). Research and Development in the Pharmaceutical Industry. Congressional Budget Office. www.cbo.gov/publication/57025

Beck, J. T., Rammage, M., Jackson, G. P., Preininger, A. M., Dankwa-Mullan, I., Roebuck, M. C., Torres, A., Holtzen, H., Coverdill, S. E., Williamson, M. P., Chau, Q., Rhee, K., & Vinegra, M. (2020). Artificial Intelligence Tool for Optimizing Eligibility Screening for Clinical Trials in a Large Community Cancer Center. *JCO Clinical Cancer Informatics*, *4*, 50–59. https://doi.org/10.1200/CCI.19.00079

Bengio, Y., Ducharme, R., Vincent, P., & Jauvin, C. (2003). A Neural Probabilistic

    Language Model. *Journal of Machine Learning Language 3.* 1137-1155.

    https://www.jmlr.org/papers/volume3/bengio03a/bengio03a.pdf

Berger, J. R., Choi, D., Kaminski, H. J., Gordon, M. F., Hurko, O., D'Cruz, O., Pleasure,

    S. J., & Feldman, E. L. (2013). Importance and hurdles to drug discovery for

    neurological disease. *Annals of Neurology*, *74*(3), 441–446.

    https://doi.org/10.1002/ana.23997

Chen, Q., Hu, Y., Peng, X., Xie, Q., Jin, Q., Gilson, A., Singer, M. B., Ai, X., Lai, P.-T.,

    Wang, Z., Keloth, V. K., Raja, K., Huang, J., He, H., Lin, F., Du, J., Zhang, R.,

    Zheng, W. J., Adelman, R. A., … Xu, H. (2023). *A systematic evaluation of large*

    *language models for biomedical natural language processing: Benchmarks,*

    *baselines, and recommendations* (Version 4). arXiv.

    https://doi.org/10.48550/ARXIV.2305.16326

Choi, J. W. (2024). *Analysis of Clinical Trial Design and Prediction of Success*.

    [Doctoral dissertation, University College London]. UCL Discovery.

ClinicalTrials.gov. (n.d.) *National Library of Medicine*. National Institutes of Health.

    Retrieved September 3, 2024, from https://clinicaltrials.gov/

Cook, J., & Ramadas, V. (2020). When to consult precision-recall curves. *The Stata*

    *Journal: Promoting Communications on Statistics and Stata*, *20*(1), 131–148.

    https://doi.org/10.1177/1536867X20909693

Dallow, N., Best, N., & Montague, T. H. (2018). Better decision making in drug

    development through adoption of formal prior elicitation. *Pharmaceutical*

    *Statistics*, *17*(4), 301–316. https://doi.org/10.1002/pst.1854

Devlin, J., Chang, M.-W., Lee, K., & Toutanova, K. (2018). *BERT: Pre-training of Deep Bidirectional Transformers for Language Understanding* (Version 2). arXiv. https://doi.org/10.48550/ARXIV.1810.04805

DiMasi, J. A., Grabowski, H. G., & Hansen, R. W. (2016). Innovation in the pharmaceutical industry: New estimates of R&D costs. *Journal of Health Economics*, *47*, 20–33. https://doi.org/10.1016/j.jhealeco.2016.01.012

Dong, X. (2018). Current Strategies for Brain Drug Delivery. *Theranostics*, *8*(6), 1481–1493. https://doi.org/10.7150/thno.21254

European Food Safety Authority. (2014). Guidance on Expert Knowledge Elicitation in Food and Feed Safety Risk Assessment. *EFSA Journal*, *12*(6). https://doi.org/10.2903/j.efsa.2014.3734

Feijoo, F., Palopoli, M., Bernstein, J., Siddiqui, S., & Albright, T. E. (2020). Key indicators of phase transition for clinical trials through machine learning. *Drug Discovery Today*, *25*(2), 414–421. https://doi.org/10.1016/j.drudis.2019.12.014

Fu, T., Huang, K., Xiao, C., Glass, L. M., & Sun, J. (2022). HINT: Hierarchical interaction network for clinical-trial-outcome predictions. *Patterns (New York, N.Y.)*, *3*(4), 100445. https://doi.org/10.1016/j.patter.2022.100445

Gao, T., Dontcheva, M., Adar, E., Liu, Z., & Karahalios, K. G. (2015). DataTone: Managing Ambiguity in Natural Language Interfaces for Data Visualization. *Proceedings of the 28th Annual ACM Symposium on User Interface Software & Technology*, 489–500. https://doi.org/10.1145/2807442.2807478

Gao, C., Fu, T., & Sun, J. (2024). *Language Interaction Network for Clinical Trial Approval Estimation* (Version 1). arXiv. https://doi.org/10.48550/ARXIV.2405.06662

Gao, C., Pradeepkumar, J., Das, T., Thati, S., & Sun, J. (2024). *Automatically Labeling $200B Life-Saving Datasets: A Large Clinical Trial Outcome Benchmark* (No. arXiv:2406.10292). arXiv. http://arxiv.org/abs/2406.10292

García, V., Mollineda, R. A., & Sánchez, J. S. (2009). Index of Balanced Accuracy: A Performance Measure for Skewed Class Distributions. In H. Araujo, A. M. Mendonça, A. J. Pinho, & M. I. Torres (Eds.), *Pattern Recognition and Image Analysis* (Vol. 5524, pp. 441–448). Springer Berlin Heidelberg. https://doi.org/10.1007/978-3-642-02172-5_57

Gayvert, K. M., Madhukar, N. S., & Elemento, O. (2016). A Data-Driven Approach to Predicting Successes and Failures of Clinical Trials. *Cell Chemical Biology*, *23*(10), 1294–1301. https://doi.org/10.1016/j.chembiol.2016.07.023

Geerts, H., Gieschke, R., & Peck, R. (2018). Use of quantitative clinical pharmacology to improve early clinical development success in neurodegenerative diseases. *Expert Review of Clinical Pharmacology*, *11*(8), 789–795. https://doi.org/10.1080/17512433.2018.1501555

Grandini, M., Bagli, E., & Visani, G. (2020). *Metrics for Multi-Class Classification: An Overview* (Version 1). arXiv. https://doi.org/10.48550/ARXIV.2008.05756

Grudzinskas, C., Dyszel, M., Sharma, K., & Gombar, C. T. (2022). Portfolio and project planning and management in the drug discovery, evaluation, development, and regulatory review process. In *Atkinson's Principles of Clinical Pharmacology* (pp. 537-562). Academic Press.

Gyawali, B., Eisenhauer, E., Tregear, M., & Booth, C. M. (2022). Progression-free survival: It is time for a new name. The Lancet Oncology, 23(3), 328–330. https://doi.org/10.1016/S1470-2045(22)00015-8

Haddad, T., Helgeson, J. M., Pomerleau, K. E., Preininger, A. M., Roebuck, M. C., Dankwa-Mullan, I., Jackson, G. P., & Goetz, M. P. (2021). Accuracy of an Artificial Intelligence System for Cancer Clinical Trial Eligibility Screening: Retrospective Pilot Study. *JMIR Medical Informatics*, *9*(3), e27767. https://doi.org/10.2196/27767

Hampson, L. V., Holzhauer, B., Bornkamp, B., Kahn, J., Lange, M. R., Luo, W., Singh, P., Ballerstedt, S., & Cioppa, G. D. (2022). A New Comprehensive Approach to Assess the Probability of Success of Development Programs Before Pivotal Trials. *Clinical Pharmacology & Therapeutics*, *111*(5), 1050–1060. https://doi.org/10.1002/cpt.2488

Hampson, L. V., Bornkamp, B., Holzhauer, B., Kahn, J., Lange, M. R., Luo, W.-L., Cioppa, G. D., Stott, K., & Ballerstedt, S. (2021). *Improving the assessment of the probability of success in late stage drug development* (No. arXiv:2102.02752). arXiv. http://arxiv.org/abs/2102.02752

Hand, D. J., Christen, P., & Kirielle, N. (2021). F*: An interpretable transformation of the F-measure. Machine Learning, 110(3), 451–456. https://doi.org/10.1007/s10994-021-05964-1

Harpum, P. (2010). *Portfolio, program, and project management in the pharmaceutical and biotechnology industries*. John Wiley & Sons. https://doi.org/10.1002/9780470603789

Holzhauer, B., Hampson, L. V., Gosling, J. P., Bornkamp, B., Kahn, J., Lange, M. R., Luo, W., Brindicci, C., Lawrence, D., Ballerstedt, S., & O'Hagan, A. (2022). Eliciting judgements about dependent quantities of interest: The SHeffield ELicitation Framework extension and copula methods illustrated using an asthma case study. *Pharmaceutical Statistics*, *21*(5), 1005–1021. https://doi.org/10.1002/pst.2212

Hossain, E., Rana, R., Higgins, N., Soar, J., Barua, P. D., Pisani, A. R., & Turner, K. (2023). Natural Language Processing in Electronic Health Records in relation to healthcare decision-making: A systematic review. *Computers in Biology and Medicine*, *155*, 106649. https://doi.org/10.1016/j.compbiomed.2023.106649

Khurana, D., Koli, A., Khatter, K., & Singh, S. (2023). Natural language processing: State of the art, current trends and challenges. *Multimedia Tools and Applications*, *82*(3), 3713–3744. https://doi.org/10.1007/s11042-022-13428-4

Kolluri, S., Lin, J., Liu, R., Zhang, Y., & Zhang, W. (2022). Machine Learning and Artificial Intelligence in Pharmaceutical Research and Development: A Review. *The AAPS Journal*, *24*(1), 19. https://doi.org/10.1208/s12248-021-00644-3

Li, B., Shin, H., Gulbekyan, G., Pustovalova, O., Nikolsky, Y., Hope, A., Bessarabova, M., Schu, M., Kolpakova-Hart, E., Merberg, D., Dorner, A., & Trepicchio, W. L. (2015). Development of a Drug-Response Modeling Framework to Identify Cell Line Derived Translational Biomarkers That Can Predict Treatment Outcome to

Erlotinib or Sorafenib. *PLOS ONE*, *10*(6), e0130700. https://doi.org/10.1371/journal.pone.0130700

Li, M., Liu, R., Lin, J., Bunn, V., & Zhao, H. (2020). Bayesian Semi-parametric Design (BSD) for adaptive dose-finding with multiple strata. *Journal of Biopharmaceutical Statistics*, *30*(5), 806–820. https://doi.org/10.1080/10543406.2020.1730870

Li, X., Lei, Y., & Ji, S. (2022). BERT- and BiLSTM-Based Sentiment Analysis of Online Chinese Buzzwords. *Future Internet*, *14*(11), 332. https://doi.org/10.3390/fi14110332

Liu, R., Lin, J., & Li, P. (2020). Design considerations for phase I/II dose finding clinical trials in Immuno-oncology and cell therapy. *Contemporary Clinical Trials*, *96*, 106083. https://doi.org/10.1016/j.cct.2020.106083

Lo, A. W., Siah, K. W., & Wong, C. H. (2017). Machine-Learning Models for Predicting Drug Approvals and Clinical-Phase Transitions. *SSRN Electronic Journal*. https://doi.org/10.2139/ssrn.2973611

Lu, Y., Chen, T., Hao, N., Van Rechem, C., Chen, J., & Fu, T. (2024). Uncertainty Quantification and Interpretability for Clinical Trial Approval Prediction. *Health Data Science*, *4*, 0126. https://doi.org/10.34133/hds.0126

Luitse, D., & Denkena, W. (2021). The great Transformer: Examining the role of large language models in the political economy of AI. *Big Data & Society*, *8*(2), 20539517211047734. https://doi.org/10.1177/20539517211047734

Mosca, E., Szigeti, F., Tragianni, S., Gallagher, D., & Groh, G. (2022). SHAP-Based Explanation Methods: A Review for NLP Interpretability. In N. Calzolari, C.-R.

Huang, H. Kim, J. Pustejovsky, L. Wanner, K.-S. Choi, … S.-H. Na (Eds.), Proceedings of the 29th International Conference on Computational Linguistics (pp. 4593–4603). Retrieved from https://aclanthology.org/2022.coling-1.406/

Murali, V., Muralidhar, Y. P., Königs, C., Nair, M., Madhu, S., Nedungadi, P., Srinivasa, G., & Athri, P. (2022). Predicting clinical trial outcomes using drug bioactivities through graph database integration and machine learning. *Chemical Biology & Drug Design*, *100*(2), 169–184. https://doi.org/10.1111/cbdd.14092

Nahm, F. S. (2022). Receiver operating characteristic curve: Overview and practical use for clinicians. *Korean Journal of Anesthesiology*, *75*(1), 25–36. https://doi.org/10.4097/kja.21209

Naidu, G., Zuva, T., & Sibanda, E. M. (2023). A Review of Evaluation Metrics in Machine Learning Algorithms. In R. Silhavy & P. Silhavy (Eds.), Artificial Intelligence Application in Networks and Systems (Vol. 724, pp. 15–25). Springer International Publishing. https://doi.org/10.1007/978-3-031-35314-7_2

Naveed, H., Khan, A. U., Qiu, S., Saqib, M., Anwar, S., Usman, M., Akhtar, N., Barnes, N., & Mian, A. (2024). *A Comprehensive Overview of Large Language Models* (No. arXiv:2307.06435). arXiv. http://arxiv.org/abs/2307.06435

Oakley, J., & O'Hagan, T. (2022). The Sheffield Elicitation Framework (SHELF). https://shelf.sites.sheffield.ac.uk/

Pappas, N., & Meyer, T. (2012). *A Survey on Language Modeling Using Neural Networks*. IDIAP Research Institute. https://infoscience.epfl.ch/server/api/core/bitstreams/066718a4-2e03-4697-a1e0-623e185d95ee/content

Peck, R. (2017). The pharmaceutical industry needs more clinical pharmacologists. British Journal of Clinical Pharmacology, 83(11), 2343–2346. https://doi.org/10.1111/bcp.13370

Paul, S. M., Mytelka, D. S., Dunwiddie, C. T., Persinger, C. C., Munos, B. H., Lindborg, S. R., & Schacht, A. L. (2010). How to improve R&D productivity: The pharmaceutical industry's grand challenge. Nature Reviews Drug Discovery, 9(3), 203–214. https://doi.org/10.1038/nrd3078

Pharmapremia Pharma Intelligence (n.d.) *Sample Selection*. Citeline. Retrieved March 3, 2025, from https://www.pharmapremiasolutions.com/table

Raiaan, M. A. K., Mukta, Md. S. H., Fatema, K., Fahad, N. M., Sakib, S., Mim, M. M. J., Ahmad, J., Ali, M. E., & Azam, S. (2024). A Review on Large Language Models: Architectures, Applications, Taxonomies, Open Issues and Challenges. *IEEE Access*, *12*, 26839–26874. https://doi.org/10.1109/ACCESS.2024.3365742

Reinisch, M., He, J., Liao, C., Siddiqui, S. A., & Xiao, B. (2024). *CTP-LLM: Clinical Trial Phase Transition Prediction Using Large Language Models* (No. arXiv:2408.10995). arXiv. http://arxiv.org/abs/2408.10995

Stalder, J. P. (2022). Creating a Clinical Development Plan. In *Project Management for Drug Developers* (pp. 185-201). CRC Press.

Staszak, M., Staszak, K., Wieszczycka, K., Bajek, A., Roszkowski, K., & Tylkowski, B. (2022). Machine learning in drug design: Use of artificial intelligence to explore the chemical structure–biological activity relationship. *WIREs Computational Molecular Science*, *12*(2), e1568. https://doi.org/10.1002/wcms.1568

Trotti, A., Colevas, A., Setser, A., Rusch, V., Jaques, D., Budach, V., Langer, C., Murphy, B., Cumberlin, R., & Coleman, C. (2003). CTCAE v3.0: Development of a comprehensive grading system for the adverse effects of cancer treatment. Seminars in Radiation Oncology, 13(3), 176–181. https://doi.org/10.1016/S1053-4296(03)00031-6

Umber, A., & Bajwa, I. S. (2011). Minimizing ambiguity in natural language software requirements specification. *2011 Sixth International Conference on Digital Information Management*, 102–107. https://doi.org/10.1109/ICDIM.2011.6093363

Vaswani, A., Shazeer, N., Parmar, N., Uszkoreit, J., Jones, L., Gomez, A. N., Kaiser, L., & Polosukhin, I. (2017). *Attention Is All You Need* (Version 7). arXiv. https://doi.org/10.48550/ARXIV.1706.03762

Vergetis, V., Skaltsas, D., Gorgoulis, V. G., & Tsirigos, A. (2021). Assessing Drug Development Risk Using Big Data and Machine Learning. *Cancer Research*, *81*(4), 816–819. https://doi.org/10.1158/0008-5472.CAN-20-0866

VeriSIM Life. (2024, August 6). How AI Assists Small Molecule Drug Design and Development. November 11, 2024, https://www.verisimlife.com/publications-blog/how-ai-assists-small-molecule-drug-design-and-development

Wallach, I., Dzamba, M., & Heifets, A. (2015). *AtomNet: A Deep Convolutional Neural Network for Bioactivity Prediction in Structure-based Drug Discovery* (Version 1). arXiv. https://doi.org/10.48550/ARXIV.1510.02855

Weissler, E. H., Naumann, T., Andersson, T., Ranganath, R., Elemento, O., Luo, Y., Freitag, D. F., Benoit, J., Hughes, M. C., Khan, F., Slater, P., Shameer, K., Roe,

M., Hutchison, E., Kollins, S. H., Broedl, U., Meng, Z., Wong, J. L., Curtis, L., … Ghassemi, M. (2021). The role of machine learning in clinical research: Transforming the future of evidence generation. *Trials*, *22*(1), 537. https://doi.org/10.1186/s13063-021-05489-x

Wilimitis, D., & Walsh, C. G. (2023). Practical Considerations and Applied Examples of Cross-Validation for Model Development and Evaluation in Health Care: Tutorial. *JMIR AI*, *2*, e49023. https://doi.org/10.2196/49023

Williamson, D. J., Struyven, R. R., Antaki, F., Chia, M. A., Wagner, S. K., Jhingan, M., Wu, Z., Guymer, R., Skene, S. S., Tammuz, N., Thomson, B., Chopra, R., & Keane, P. A. (2024). Artificial Intelligence to Facilitate Clinical Trial Recruitment in Age-Related Macular Degeneration. *Ophthalmology Science*, *4*(6), 100566. https://doi.org/10.1016/j.xops.2024.100566

Wong, C. H., Siah, K. W., & Lo, A. W. (2019). Estimation of clinical trial success rates and related parameters. *Biostatistics*, *20*(2), 273–286. https://doi.org/10.1093/biostatistics/kxx069

Wouters, O. J., McKee, M., & Luyten, J. (2020). Estimated research and development investment needed to bring a new medicine to market, 2009-2018. *JAMA*, 323(9), 844-853. https://doi.org/10.1001/jama.2020.1166

Wouters, O. J., & Kesselheim, A. S. (2024). Quantifying Research and Development Expenditures in the Drug Industry. JAMA Network Open, 7(6), e2415407. https://doi.org/10.1001/jamanetworkopen.2024.15407

Yamaguchi, S., Kaneko, M., & Narukawa, M. (2021). Approval success rates of drug candidates based on target, action, modality, application, and their combinations.

*Clinical and Translational Science*, *14*(3), 1113–1122.

https://doi.org/10.1111/cts.12980

Zaragoza Domingo, S., Alonso, J., Ferrer, M., Acosta, M. T., Alphs, L., Annas, P., Balabanov, P., Berger, A.-K., Bishop, K. I., Butlen-Ducuing, F., Dorffner, G., Edgar, C., De Gracia Blanco, M., Harel, B., Harrison, J., Horan, W. P., Jaeger, J., Kottner, J., Pinkham, A., … Yavorsky, C. (2024). Methods for Neuroscience Drug Development: Guidance on Standardization of the Process for Defining Clinical Outcome Strategies in Clinical Trials. *European Neuropsychopharmacology*, *83*, 32–42. https://doi.org/10.1016/j.euroneuro.2024.02.009

Zheng, W., Peng, D., Xu, H., Li, Y., Zhu, H., Fu, T., & Yao, H. (2024). *Multimodal Clinical Trial Outcome Prediction with Large Language Models* (Version 3). arXiv. https://doi.org/10.48550/ARXIV.2402.06512

Zhou, S., & Johnson, R. (2018). *Pharmaceutical Probability of Success*. Alacrita. https://alacrita.com/wp-content/uploads/2018/12/Pharmaceutical-Probability-of-Success.pdf

Zhu, K., Zheng, Y., & Chan, K. C. G. (2024). *Weighted Brier Score—An Overall Summary Measure for Risk Prediction Models with Clinical Utility Consideration* (No. arXiv:2408.01626). arXiv. https://doi.org/10.48550/arXiv.2408.01626

Zhu, T. (2021). Challenges of Psychiatry Drug Development and the Role of Human Pharmacology Models in Early Development—A Drug Developer's Perspective. *Frontiers in Psychiatry*, *11*, 562660. https://doi.org/10.3389/fpsyt.2020.562660

# Appendix A

## Table A-1. Full Logistic Regression Results

| Training/Testing | ROC-AUC | TN | FP | FN | TP | Accuracy | Bal'd Acc | Precision | Recall | F1 | Cohen-k | PR-AUC |
|---|---|---|---|---|---|---|---|---|---|---|---|---|
| All Indications Phase 1 | 0.665 | 1174 | 862 | 3665 | 7636 | 0.661 | 0.626 | 0.899 | 0.676 | 0.771 | 0.161 | 0.905 |
| All Indications Phase 2 | 0.602 | 1672 | 500 | 5385 | 2972 | 0.441 | 0.563 | 0.856 | 0.356 | 0.502 | 0.0685 | 0.844 |
| All Indications Phase 3 | 0.661 | 660 | 341 | 2028 | 2739 | 0.589 | 0.617 | 0.889 | 0.575 | 0.698 | 0.14 | 0.896 |
| Neuroscience Only Phase 1 | 0.652 | 522 | 261 | 1662 | 2024 | 0.57 | 0.607 | 0.886 | 0.549 | 0.678 | 0.127 | 0.887 |
| Neuroscience Only Phase 2 | 0.593 | 673 | 134 | 2013 | 779 | 0.403 | 0.556 | 0.853 | 0.279 | 0.421 | 0.062 | 0.831 |
| Neuroscience Only Phase 3 | 0.66 | 268 | 113 | 695 | 815 | 0.573 | 0.621 | 0.878 | 0.54 | 0.669 | 0.155 | 0.885 |
| Mean | 0.639 | 828 | 369 | 2575 | 2828 | 0.540 | 0.598 | 0.877 | 0.496 | 0.623 | 0.119 | 0.875 |

TN = true negative. FP = false positive. FN = false negative. TP = true positive.

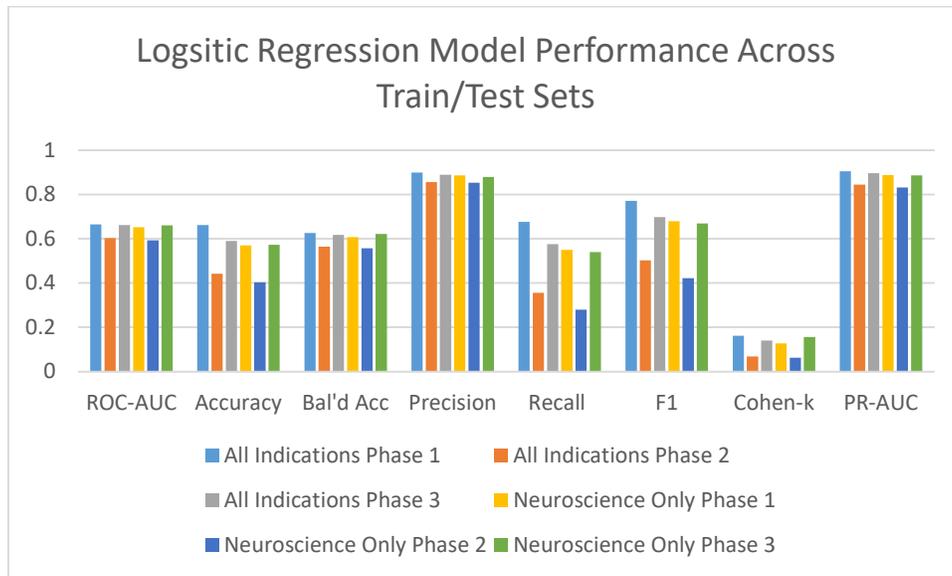

**Figure A-1. Full Logistic Regression Results**

## Table A-2. Full Gradient Boosting Results

| Training/Testing | ROC-AUC | TN | FP | FN | TP | Accuracy | Bal'd Acc | Precision | Recall | F1 | Cohen-k | PR-AUC |
|---|---|---|---|---|---|---|---|---|---|---|---|---|
| All Indications Phase 1 | 0.67 | 926 | 1110 | 2683 | 8618 | 0.716 | 0.609 | 0.886 | 0.763 | 0.82 | 0.165 | 0.909 |
| All Indications Phase 2 | 0.598 | 1491 | 681 | 4567 | 3790 | 0.501 | 0.57 | 0.848 | 0.454 | 0.591 | 0.0842 | 0.848 |
| All Indications Phase 3 | 0.662 | 583 | 418 | 1568 | 3199 | 0.656 | 0.627 | 0.884 | 0.671 | 0.763 | 0.174 | 0.893 |
| Neuroscience Only Phase 1 | 0.662 | 446 | 337 | 1289 | 2397 | 0.636 | 0.61 | 0.877 | 0.65 | 0.747 | 0.149 | 0.895 |
| Neuroscience Only Phase 2 | 0.591 | 597 | 210 | 1716 | 1076 | 0.465 | 0.563 | 0.837 | 0.385 | 0.528 | 0.075 | 0.835 |
| Neuroscience Only Phase 3 | 0.652 | 208 | 173 | 521 | 989 | 0.633 | 0.6 | 0.851 | 0.655 | 0.74 | 0.15 | 0.878 |
| Mean | 0.639 | 708.500 | 488.167 | 2057.333 | 3344.833 | 0.601 | 0.597 | 0.864 | 0.596 | 0.698 | 0.133 | 0.876 |

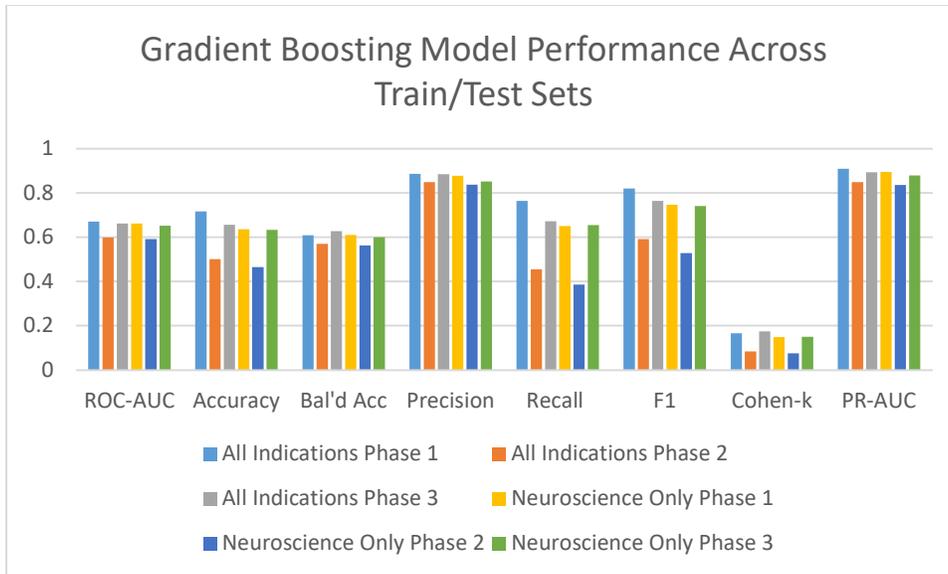

Figure A-2. Full Gradient Boosting Results

Table A-3. Full Random Forest Results

| Training/Testing | ROC-AUC | TN | FP | FN | TP | Accuracy | Bal'd Acc | Precision | Recall | F1 | Cohen-k | PR-AUC |
|---|---|---|---|---|---|---|---|---|---|---|---|---|
| All Indications Phase 1 | 0.664 | 957 | 1079 | 2830 | 8471 | 0.707 | 0.61 | 0.887 | 0.75 | 0.813 | 0.162 | 0.907 |
| All Indications Phase 2 | 0.599 | 1493 | 679 | 4596 | 3761 | 0.499 | 0.569 | 0.847 | 0.45 | 0.588 | 0.0824 | 0.847 |
| All Indications Phase 3 | 0.651 | 586 | 415 | 1743 | 3024 | 0.626 | 0.609 | 0.879 | 0.634 | 0.737 | 0.144 | 0.886 |
| Neuroscience Only Phase 1 | 0.658 | 475 | 308 | 1400 | 2286 | 0.618 | 0.613 | 0.881 | 0.62 | 0.728 | 0.146 | 0.893 |
| Neuroscience Only Phase 2 | 0.588 | 600 | 207 | 1734 | 1058 | 0.461 | 0.561 | 0.836 | 0.379 | 0.522 | 0.0732 | 0.825 |
| Neuroscience Only Phase 3 | 0.657 | 242 | 139 | 627 | 883 | 0.595 | 0.61 | 0.864 | 0.585 | 0.697 | 0.149 | 0.88 |
| Mean | 0.636 | 726 | 471 | 2155 | 3247 | 0.584 | 0.595 | 0.866 | 0.570 | 0.681 | 0.126 | 0.873 |

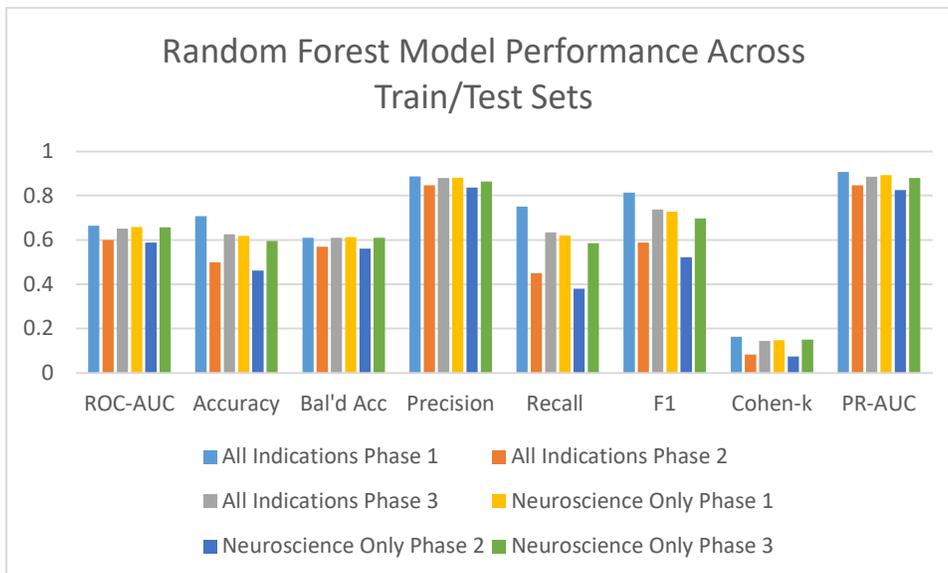

Figure A-3. Full Random Forest Results

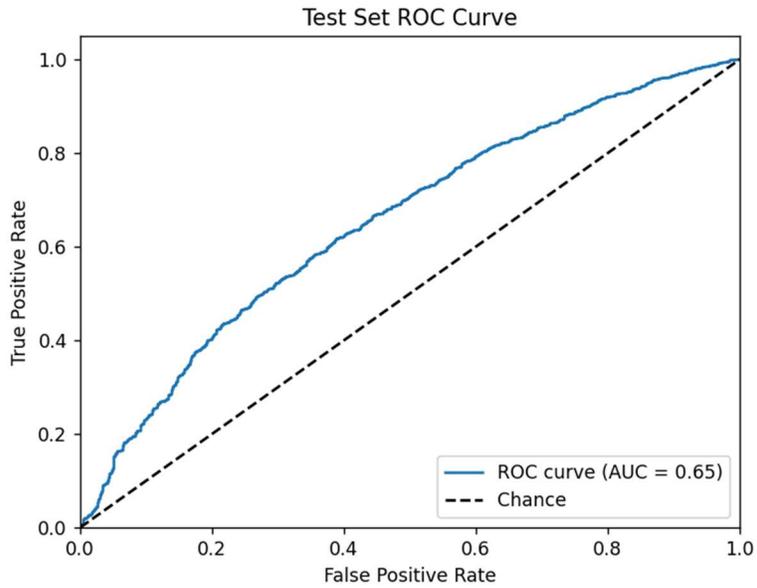

**Figure A4. ROC Curve for Phase 1 Logistic Regression Model Training on Neuroscience Trials**

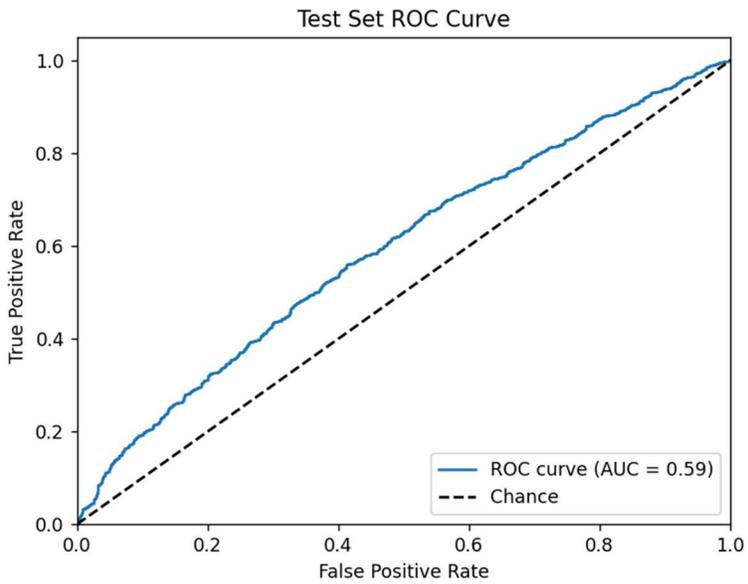

**Figure A5. ROC Curve for Phase 2 Logistic Regression Model Training on Neuroscience Trials**

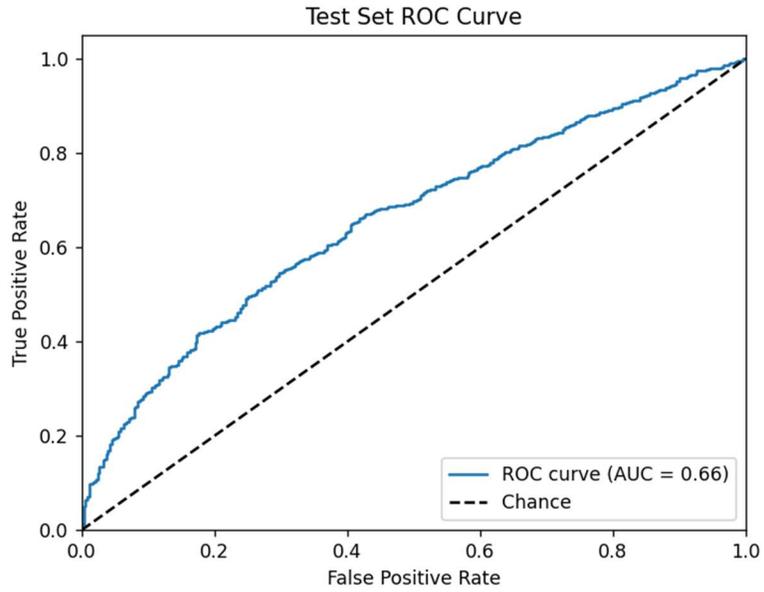

**Figure A6. ROC Curve for Phase 3 Logistic Regression Model Training on Neuroscience Trials**

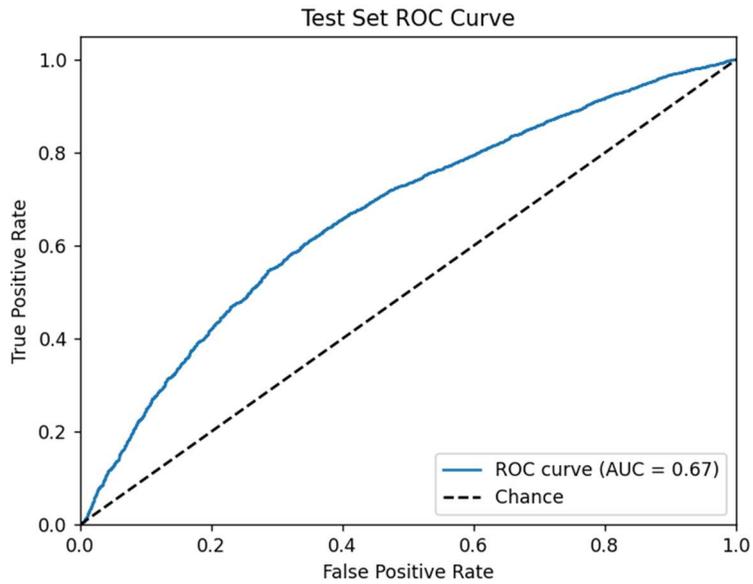

**Figure A7. ROC Curve for Phase 1 Logistic Regression Model Training on All Indications**

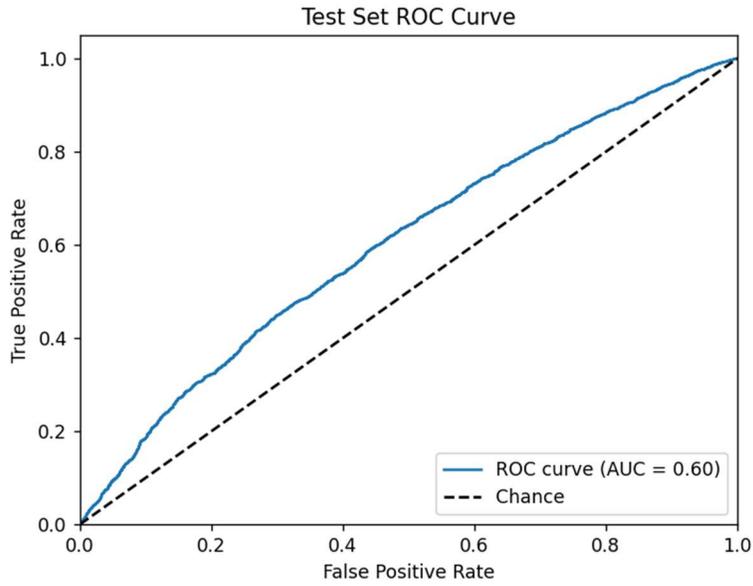

**Figure A8. ROC Curve for Phase 2 Logistic Regression Model Training on All Indications**

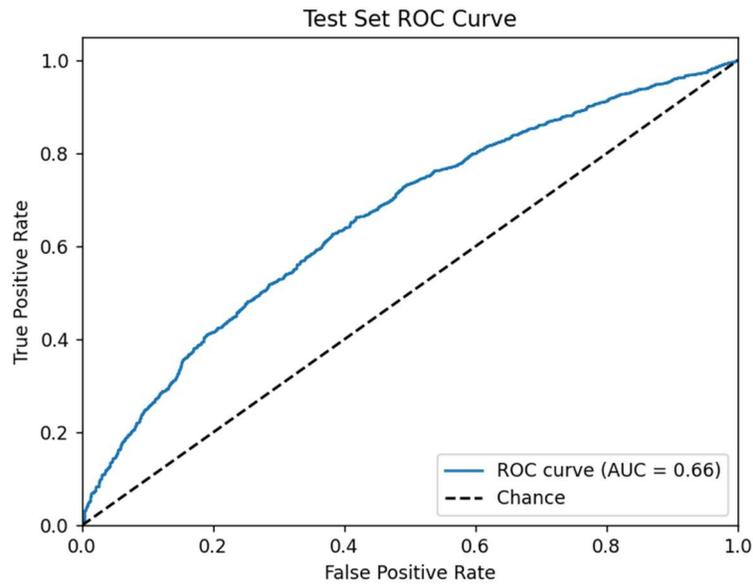

**Figure A9. ROC Curve for Phase 3 Logistic Regression Model Training on All Indications**

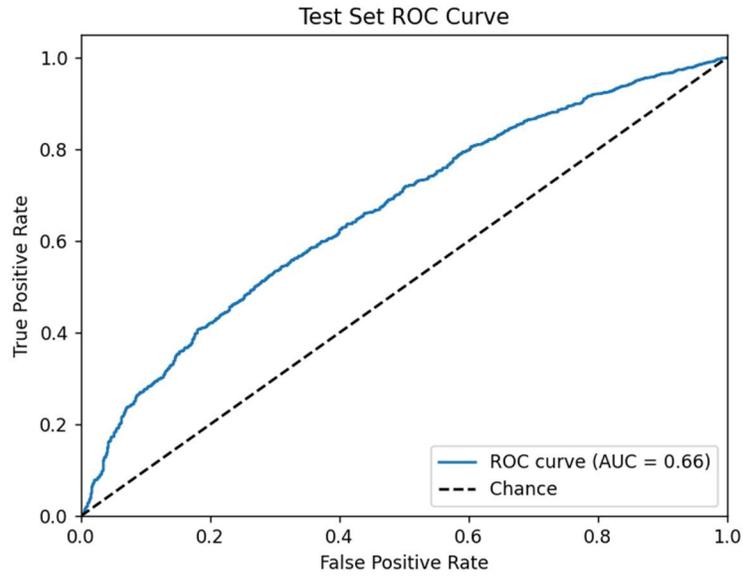

**Figure A10. ROC Curve for Phase 1 Gradient Boosting Model Training on Neuroscience Trials**

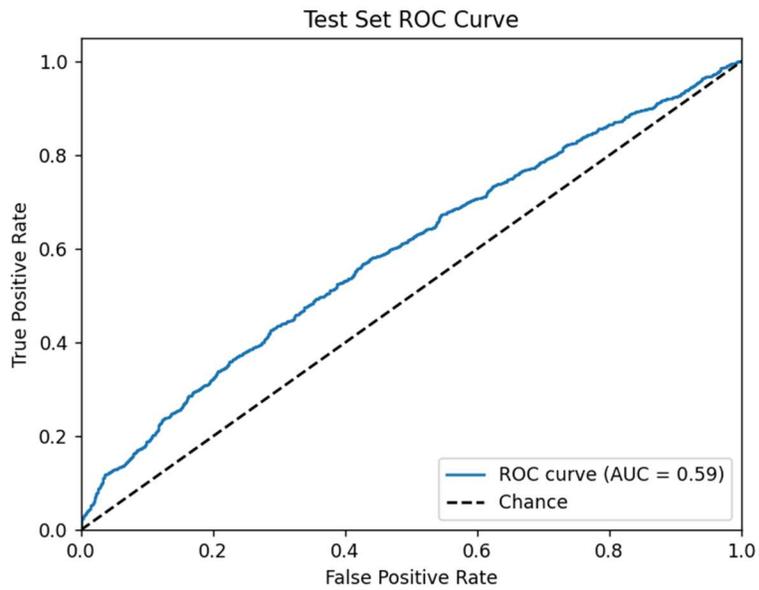

**Figure A11. ROC Curve for Phase 2 Gradient Boosting Model Training on Neuroscience Trials**

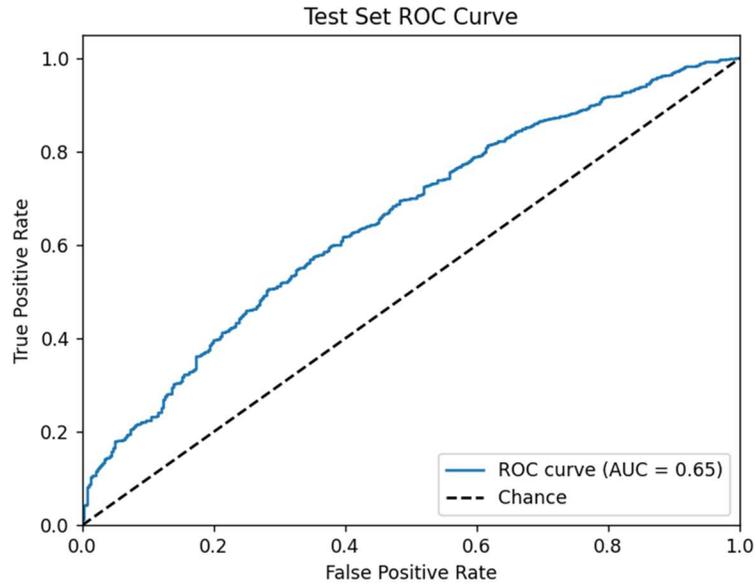

**Figure A12. ROC Curve for Phase 3 Gradient Boosting Model Training on Neuroscience Trials**

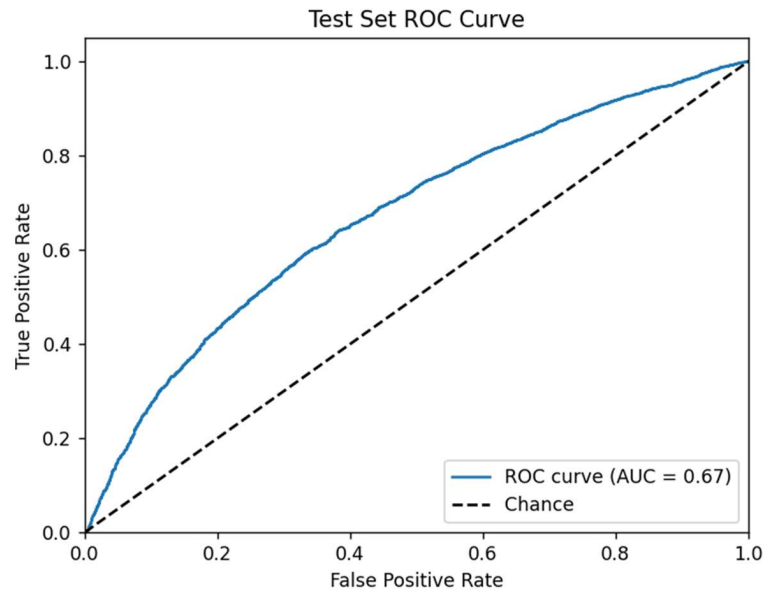

**Figure A13. ROC Curve for Phase 1 Gradient Boosting Model Training on All Indications**

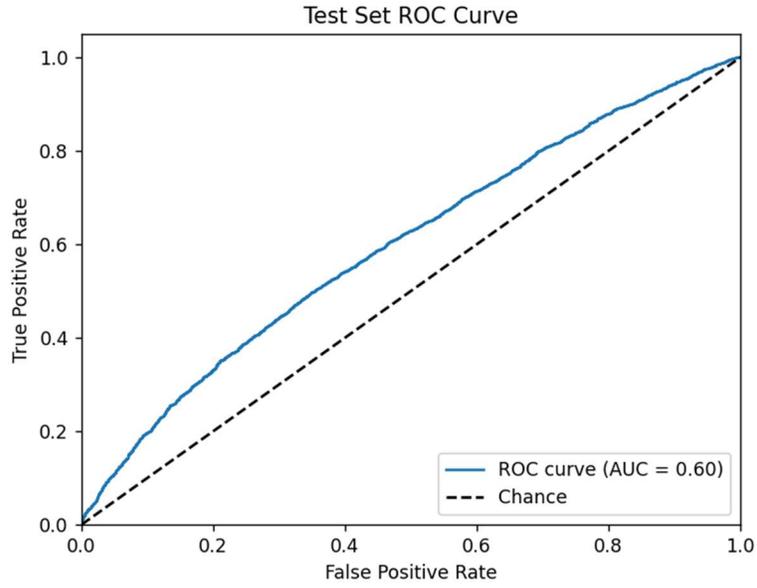

**Figure A14. ROC Curve for Phase 2 Gradient Boosting Model Training on All Indications**

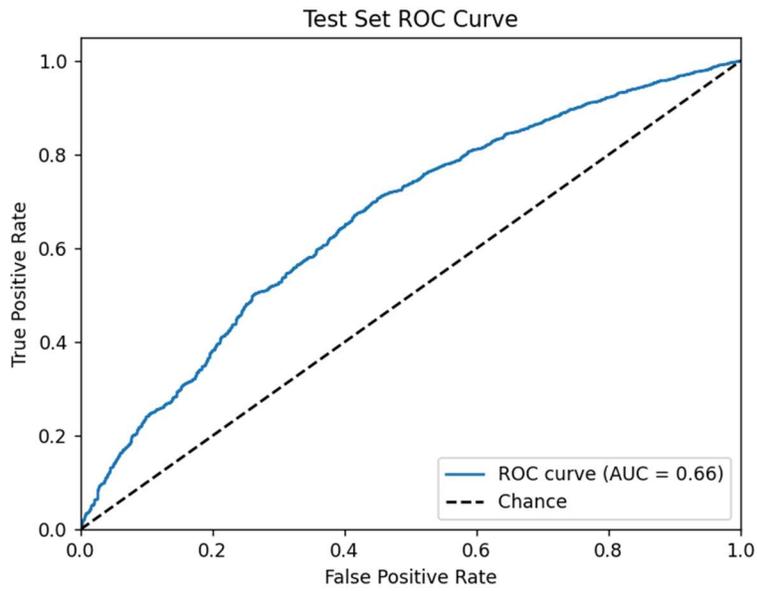

**Figure A15. ROC Curve for Phase 3 Gradient Boosting Model Training on All Indications**

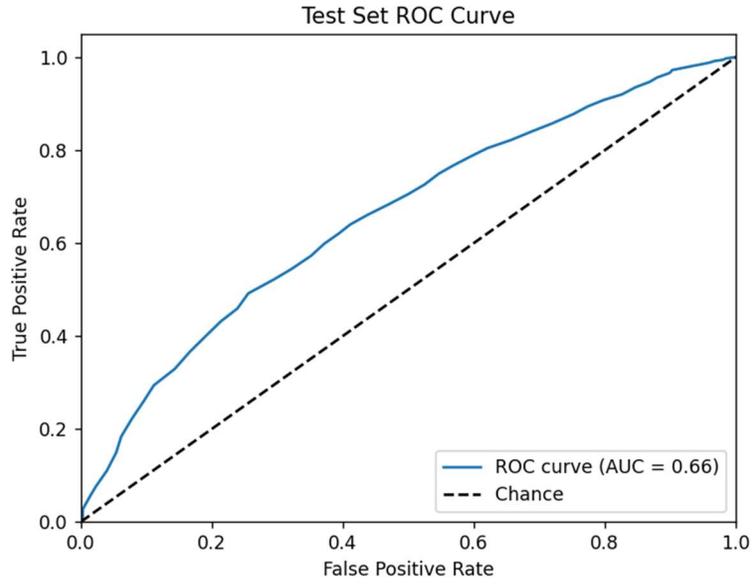

**Figure A16. ROC Curve for Phase 1 Random Forest Model Training on Neuroscience Trials**

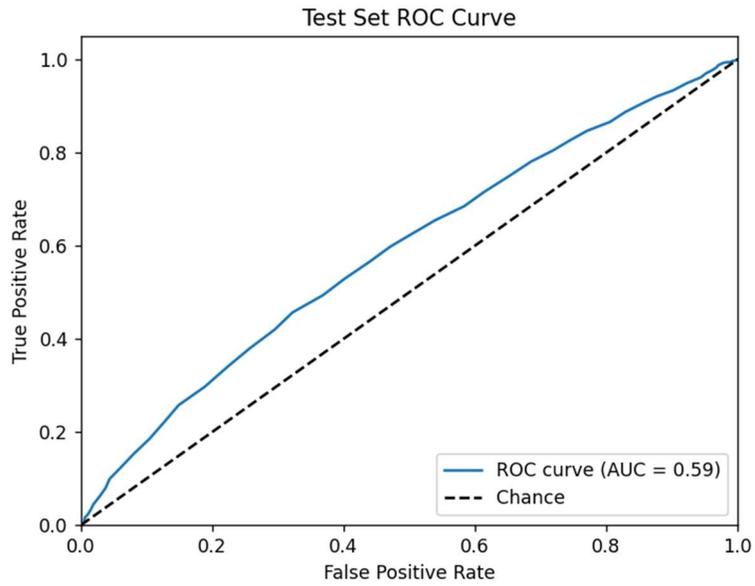

**Figure A17. ROC Curve for Phase 2 Random Forest Model Training on Neuroscience Trials**

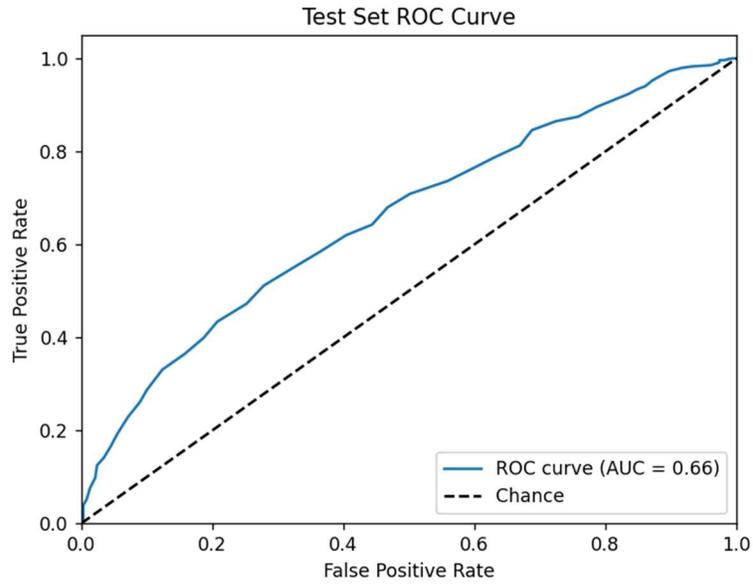

**Figure A18. ROC Curve for Phase 3 Random Forest Model Training on Neuroscience Trials**

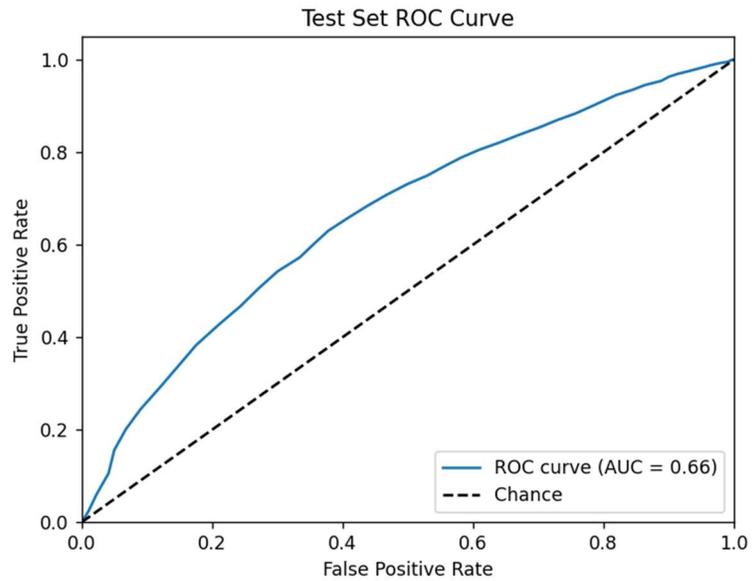

**Figure A19. ROC Curve for Phase 1 Random Forest Model Training on All Indications**

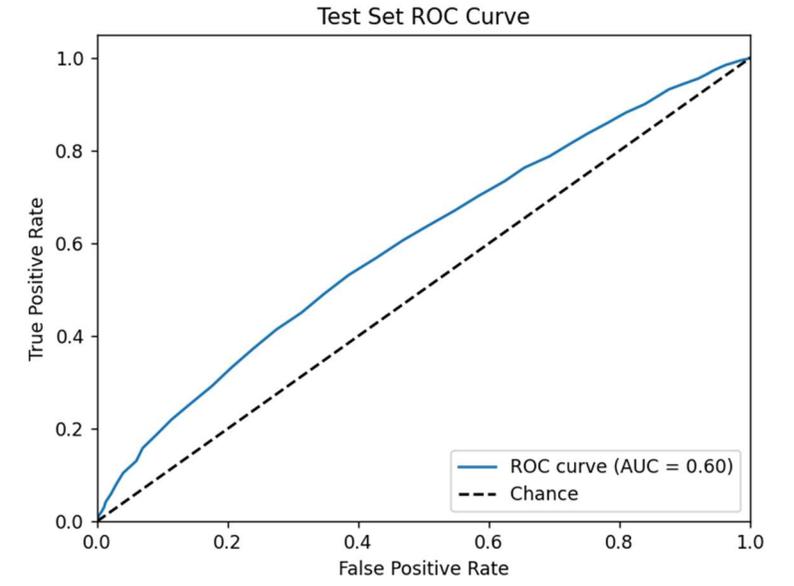

**Figure A20. ROC Curve for Phase 2 Random Forest Model Training on All Indications**

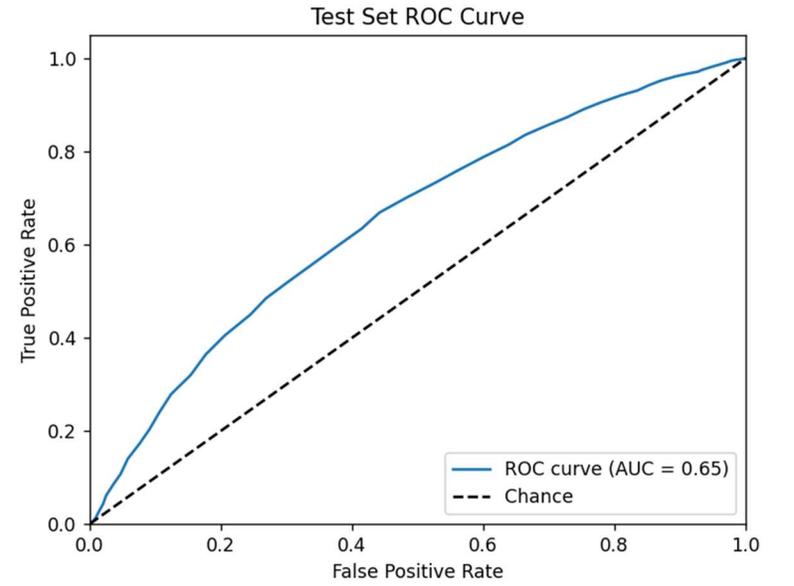

**Figure A21. ROC Curve for Phase 3 Random Forest Model Training on All Indications**

# Appendix B

**Neuroscience Filter Strings**

**Dementias and related**
"Alzheimer's disease", "Alzheimer's", "Alzheimers disease", "Alzheimers", "AD",
"Parkinson's disease", "Parkinson's", "Parkinsons disease", "Parkinsons", "PD",
"Huntington's disease", "Huntingtons disease", "Huntington", "HD",
"Amyotrophic lateral sclerosis", "ALS", "Lou Gehrig's disease",
"Frontotemporal dementia", "FTD", "fronto-temporal dementia", "frontotemporal lobar degeneration", "FTLD",
"Lewy body dementia", "LBD", "dementia with Lewy bodies", "DLB",
"Vascular dementia", "multi-infarct dementia",
"Creutzfeldt-Jakob disease", "CJD",
"Progressive supranuclear palsy", "PSP", "Steele-Richardson-Olszewski syndrome",
"Corticobasal degeneration", "CBD", "corticobasal syndrome", "CBS",

**Multiple sclerosis & related**
"Multiple sclerosis", "MS",
"Neuromyelitis optica", "NMO", "Devic's disease",
"Acute disseminated encephalomyelitis", "ADEM",
"Transverse myelitis",

**Neuropathies, demyelinating, peripheral nerve**
"Guillain-Barré syndrome", "GBS",
"Chronic inflammatory demyelinating polyneuropathy", "CIDP",
"Peripheral neuropathy", "peripheral neuropathic", "neuropathy", "neuropathic",
"Diabetic neuropathy", "diabetic neuropathic",
"Charcot-Marie-Tooth disease", "CMT", "hereditary motor and sensory neuropathy", "HMSN",
"Friedreich's ataxia", "FRDA",
"Spinocerebellar ataxia", "SCA", "ataxia", "ataxic disorders", "ataxic",

**Movement disorders**
"Essential tremor", "familial tremor",
"Dystonia", "dystonic",
"Tourette syndrome", "Tourette's", "Tourette", "TS",
"Restless legs syndrome", "RLS", "Willis-Ekbom disease",
"Multiple system atrophy", "MSA", "Shy-Drager syndrome",
"Ataxia-telangiectasia", "Louis–Bar syndrome",
"Myasthenia gravis", "MG",
"Lambert-Eaton myasthenic syndrome", "LEMS",

**Muscular dystrophies & SMA**
  "Muscular dystrophy", "MD",
  "Duchenne muscular dystrophy", "DMD",
  "Becker muscular dystrophy", "BMD",
  "Spinal muscular atrophy", "SMA",

**Myopathies, inflammatory muscle diseases**
  "Polymyositis",
  "Dermatomyositis",

**Epilepsies & seizure disorders**
  "Epilepsy", "epileptic disorders", "seizure disorder", "seizures",
  "Absence seizures", "petit mal seizures",
  "Tonic-clonic seizures", "grand mal seizures",
  "Temporal lobe epilepsy",
  "Dravet syndrome", "severe myoclonic epilepsy of infancy",
  "Lennox-Gastaut syndrome", "LGS",
  "West syndrome", "infantile spasms",
  "Landau-Kleffner syndrome",
  "Benign rolandic epilepsy", "benign childhood epilepsy with centrotemporal spikes",

**Headaches & facial pain**
  "Migraine", "migraines", "migraine headache",
  "Tension-type headache", "tension headache",
  "Cluster headache", "cluster headaches",
  "Trigeminal neuralgia", "tic douloureux",
  "Occipital neuralgia",
  "Hemicrania continua",
  "Postherpetic neuralgia", "PHN",
  "Neuropathic pain",
  "Complex regional pain syndrome", "CRPS", "reflex sympathetic dystrophy", "RSD",

**Stroke & cerebrovascular**
  "Stroke", "cerebrovascular accident", "CVA",
  "Transient ischemic attack", "TIA",
  "Subarachnoid hemorrhage", "SAH",
  "Cerebral aneurysm", "intracranial aneurysm",
  "Arteriovenous malformation", "AVM",
  "Intracerebral hemorrhage", "ICH",
  "Vascular malformation",

**Infections of the CNS**
  "Meningitis",
  "Encephalitis",
  "Brain abscess", "intracranial abscess",
  "Poliomyelitis", "polio",
  "Rabies",
  "Tetanus",
  "Progressive multifocal leukoencephalopathy", "PML",
  "Lyme neuroborreliosis", "neurological Lyme disease",
  "Herpes simplex encephalitis",
  "Prion disease", "prion disorders",

**Brain & spinal tumors**
  "Brain tumor", "intracranial neoplasm",
  "Glioblastoma", "GBM", "glioblastoma multiforme",
  "Astrocytoma",
  "Meningioma",
  "Oligodendroglioma",
  "Spinal cord tumor", "spinal neoplasm",

**Structural/anatomical brain & spine**
  "Hydrocephalus",
  "Chiari malformation", "Arnold-Chiari malformation",
  "Syringomyelia",
  "Tethered cord syndrome",

**Conversion / functional disorders**
  "Conversion disorder", "functional neurological symptom disorder",
  "Psychogenic nonepileptic seizures", "PNES",

**Autonomic dysfunction**
  "Dysautonomia", "autonomic dysfunction",
  "Postural orthostatic tachycardia syndrome", "POTS",

**Sleep disorders (neurological basis)**
  "Narcolepsy", "narcoleptic",
  "REM sleep behavior disorder", "RBD",
  "Idiopathic hypersomnia",

**Developmental & pediatric**
  "Autism spectrum disorder", "ASD", "autistic disorder",
  "Cerebral palsy", "CP",
  "Rett syndrome", "RTT",

"Fragile X syndrome",
"Angelman syndrome",
"Developmental coordination disorder", "DCD",
"Attention deficit hyperactivity disorder", "ADHD",

**Rare/Other**
"Guam disease", "lytico-bodig disease",
"Familial hemiplegic migraine", "FHM",
"CADASIL", "Cerebral autosomal dominant arteriopathy with subcortical infarcts and leukoencephalopathy",
"Moyamoya disease",
"Stiff-person syndrome", "SPS", "stiff-man syndrome",
"Opsoclonus-myoclonus syndrome", "OMS", "dancing eyes-dancing feet syndrome",
"Neuromyopathy", "neuromuscular disorder",
"Bell's palsy", "idiopathic facial nerve palsy",
"Neurofibromatosis", "NF",
"Tuberous sclerosis", "Bourneville disease", "TSC",
"Sturge-Weber syndrome", "encephalotrigeminal angiomatosis",
"Alexander disease",
"Canavan disease",
"Krabbe disease", "globoid cell leukodystrophy",
"Metachromatic leukodystrophy", "MLD",
"Adrenoleukodystrophy", "ALD",
"Werdnig-Hoffmann disease", "SMA Type I"
"Kennedy's disease", "spinal and bulbar muscular atrophy", "SBMA",
"Wilson's disease", "hepatolenticular degeneration",
"Neuroacanthocytosis", "Levine-Critchley syndrome",
"Subacute sclerosing panencephalitis", "SSPE",
"Progressive encephalopathy",
"Aphasia", "dysphasia",
"Agnosia",
"Apraxia", "dyspraxia",
"Locked-in syndrome",
"Guillain-Barré variants", "Miller Fisher syndrome", "MFS",
"Sarcoidosis", "neurosarcoidosis",
"Behçet's disease", "neuro-Behçet's",
"Bickerstaff brainstem encephalitis",
"Acoustic neuroma", "vestibular schwannoma",
"Atypical parkinsonism", "Parkinson-plus syndrome",
"Essential palatal tremor",
"Orthostatic tremor",
"Hereditary spastic paraplegia", "HSP", "familial spastic paraparesis",

"Paramyotonia congenita",
"Periodic paralysis", "hypokalemic periodic paralysis", "hyperkalemic periodic paralysis",
"Myotonic dystrophy", "DM", "Steinert's disease",
"Paraneoplastic neurological syndromes", "PNS",
"Gerstmann-Sträussler-Scheinker syndrome", "GSS",
"Leukoencephalopathy",
"Niemann-Pick disease type C",
"Tay-Sachs disease",
"Sandhoff disease",
"Neurodegeneration with brain iron accumulation", "NBIA", "Hallervorden-Spatz disease",
"Phenylketonuria", "PKU",
"Subacute combined degeneration",
"Toxic encephalopathy",
"Wernicke encephalopathy",
"Korsakoff syndrome", "Wernicke-Korsakoff syndrome",
"Marchiafava-Bignami disease",
"Hashimoto's encephalopathy",
"Posterior reversible encephalopathy syndrome", "PRES",
"Fibromyalgia",
"Chronic fatigue syndrome", "ME/CFS", "myalgic encephalomyelitis",
"Hyperekplexia", "startle disease",
"Pudendal neuropathy",
"Temporal arteritis", "giant cell arteritis",
"Reversible cerebral vasoconstriction syndrome", "RCVS",
"Carpal tunnel syndrome",
"Cubital tunnel syndrome",
"Thoracic outlet syndrome (neurogenic)",
"Meralgia paresthetica", "lateral femoral cutaneous neuropathy",
"Spinal stenosis", "neurogenic claudication",
"Radiculopathy", "radicular pain",
"Facet arthropathy",
"Syringobulbia",
"Brown-Séquard syndrome",
"Central cord syndrome",
"Anterior cord syndrome",
"Subacute sclerosing encephalitis",
"Botulism",
"Toxic neuropathy",
"Heavy metal neuropathy", "lead neuropathy", "mercury neuropathy",
"Reye's syndrome",
"Vogt-Koyanagi-Harada disease",

"Susac syndrome",
"Stargardt disease (neurological implications)",
"Usher syndrome",
"Optic neuritis",
"Alpers' disease", "Alpers' syndrome",
"Refsum disease",
"Canal dehiscence syndrome", "superior semicircular canal dehiscence",
"Barre-Lieou syndrome",
"Ehlers-Danlos syndrome (neurological complications)",
"Marfan syndrome (neurological complications)",
"Zellweger syndrome",
"Agenesis of the corpus callosum",
"Encephalocele",
"Anencephaly",
"Spina bifida",
"Subdural hematoma",
"Epidural hematoma",
"Concussion", "mild traumatic brain injury", "TBI",
"Chronic traumatic encephalopathy", "CTE",
"Diffuse axonal injury",
"Pseudotumor cerebri", "idiopathic intracranial hypertension", "IIH",
"Alice in Wonderland syndrome", "AIWS",
"Synesthesia", "synaesthesia",
"Phantom limb pain", "phantom limb syndrome",
"Fetal alcohol syndrome (neurological defects)",
"Chemotherapy-induced neuropathy",
"Radiation-induced neuropathy", "radiation encephalopathy",
"Neurotoxicity",
"Neuroleptic malignant syndrome", "NMS",
"Serotonin syndrome",
"Tardive dyskinesia",
"Akathisia",
"Antiphospholipid syndrome (neurological manifestations)",
"Fabry disease (neurological involvement)",
"Cerebrotendinous xanthomatosis", "CTX",
"MELAS (Mitochondrial encephalomyopathy, lactic acidosis, and stroke-like episodes)",
"MERRF (Myoclonic epilepsy with ragged red fibers)",
"Leigh syndrome", "subacute necrotizing encephalomyelopathy",
"Kearns-Sayre syndrome",
"SANDO (Sensory ataxic neuropathy, dysarthria, and ophthalmoparesis)",
"Wolfram syndrome", "DIDMOAD",
"Basal ganglia calcification", "Fahr's disease",

"Neurocysticercosis",
"Cerebral malaria",
"Sleeping sickness", "African trypanosomiasis",
"Kuru",
"New variant CJD", "vCJD",
"Foix-Chavany-Marie syndrome",
"Rasmussen's encephalitis",
"Hemimegalencephaly",
"Neonatal encephalopathy", "hypoxic-ischemic encephalopathy",
"Neuroblastoma (paraneoplastic)",
"Opsoclonus-myoclonus-ataxia",
"Neurodegeneration with Lewy bodies",
"Hypomyelination disorders",
"Paroxysmal kinesigenic dyskinesia", "PKD",
"Paroxysmal nonkinesigenic dyskinesia", "PNKD",
"Alternating hemiplegia of childhood",
"Abetalipoproteinemia (neurological complications)",
"Foix-Alajouanine syndrome", "subacute necrotic myelopathy",
"Hirayama disease", "monomelic amyotrophy",
"Floppy baby syndrome",
"Pompe disease", "GSD II",
"Mitochondrial myopathy",
"Glycogen storage disease type IV", "Andersen's disease",
"Hyperammonemia (neurological manifestations)",
"Maple syrup urine disease",
"Urea cycle disorders (neurological manifestations)",
"TRAPS syndrome", "Tumor necrosis factor receptor–associated periodic syndrome",
"Cogan's syndrome (neurological involvement)",
"Retinal vasculopathy with cerebral leukodystrophy", "RVCL",
"Sneddon's syndrome",
"Takayasu arteritis (neurological complications)",
"Cerebral amyloid angiopathy",
"Binswanger's disease", "subcortical leukoencephalopathy",
"Neuro-ichthyosis",
"Gaucher disease (neuronopathic forms)",
"Parkinsonism-dementia complex of Guam", "Lytico-Bodig disease",
"Farber disease (neurologic involvement)",
"Adrenomyeloneuropathy", "AMN",
"Pelizaeus-Merzbacher disease", "PMD",
"Hereditary diffuse leukoencephalopathy with axonal spheroids",
"Aicardi syndrome",
"Goldenhar syndrome (neurologic involvement)",

"Moebius syndrome (facial nerve involvement)",
"Klippel-Feil syndrome",
"CHARGE syndrome",
"Pitt-Hopkins syndrome",
"Phelan-McDermid syndrome", "22q13 deletion syndrome",
"Smith-Magenis syndrome",
"Prader-Willi syndrome",
"Velo-cardio-facial syndrome", "22q11.2 deletion syndrome",
"Williams syndrome",
"Down syndrome (neurological aspects)",
"Trisomy 18", "Edwards syndrome",
"Trisomy 13", "Patau syndrome",
"Hyperekplexia (startle disease)",
"Neonatal seizures",
"Infantile spasms (West syndrome)",
"Benign myoclonus of early infancy",
"Benign familial neonatal convulsions",
"Ohtahara syndrome", "early infantile epileptic encephalopathy",
"Doose syndrome", "myoclonic-astatic epilepsy",
"Eclampsia",
"Transient global amnesia",
"Amnestic syndromes",
"Demyelinating polyneuropathy",
"Glossopharyngeal neuralgia",
"Post-polio syndrome",
"Hereditary spastic paraplegia (repeated synonyms)",
"Motor neuron disease",
"Primary lateral sclerosis", "PLS",
"Progressive muscular atrophy", "PMA",
"Bulbar palsy", "pseudobulbar palsy",
"Polyradiculopathy",
"Cauda equina syndrome",
"Conus medullaris syndrome",
"Brain ischemia", "cerebral ischemia",
"White matter disease",
"Leukodystrophy",
"Subcortical arteriosclerotic encephalopathy",
"Neurodegenerative disease",
"Movement disorder",
"Spastic paraplegia",
"Neurodevelopmental disorder",
"Neurocutaneous syndrome",
"Inborn error of metabolism (neurological involvement)",

"Lysosomal storage disease (neurological involvement)",
"Ulnar neuropathy",
"Tarsal tunnel syndrome",
"Meralgia paresthetica (lateral femoral cutaneous neuropathy)",
"Complex migraine", "basilar-type migraine", "hemiplegic migraine",
"Basilar migraine", "basilar-type migraine",
"Ophthalmoplegic migraine",
"Retinal migraine",
"Post-traumatic headache",
"Medication overuse headache", "MOH", "rebound headache",
"Opioid-induced hyperalgesia",
"Sensorineural hearing loss (neurological basis)",
"Vertigo", "labyrinthitis", "vestibular neuritis",
"Meniere's disease", "endolymphatic hydrops",
"Cholesteatoma (intracranial extension)",
"Neuro-ophthalmologic disorders",
"Leber's hereditary optic neuropathy", "LHON",
"Batten disease", "neuronal ceroid lipofuscinosis", "NCL", "CLN1 disease", "CLN2 disease",
"Galactosemia (neurological complications)",
"Dejerine-Sottas disease", "hereditary motor and sensory neuropathy type III",
"Acute flaccid myelitis", "AFM",
"West Nile neurological syndrome",
"Zika-associated neuropathy", "microcephaly",
"COVID-19 neurological complications",
"MERS neurological complications",
"SARS neurological complications",
"HIV-associated neurocognitive disorder", "HAND",
"Neurosyphilis",
"Neuro-Behçet's disease",
"Whipple's disease (neurological)",
"Celiac disease (neurological manifestations)", "gluten ataxia", "celiac neuropathy",
"Isaacs' syndrome", "continuous muscle fiber activity syndrome",
"Neuromyotonia",
"Rippling muscle disease",
"Dermatome pain", "radicular syndrome",
"Herpes zoster", "shingles (neurological complications)",
"Von Hippel-Lindau disease", "hemangioblastomas",
"Marinesco-Sjögren syndrome",
"Cogan's syndrome",
"Neurotuberculosis", "tuberculous meningitis",
"Brucellosis (neurobrucellosis)",
"Coccidioidomycosis (neurological)",

"Histoplasmosis (neurohistoplasmosis)",
"Fatal familial insomnia", "FFI",
"Neurolymphomatosis",
"Lymphomatoid granulomatosis (CNS involvement)",
"Progressive rubella panencephalitis",
"Parry-Romberg syndrome", "hemifacial atrophy",
"Hemifacial spasm",
"Trigeminal autonomic cephalalgias", "TACs",
"SUNCT syndrome", "Short-lasting unilateral neuralgiform headache with conjunctival injection and tearing",
"SUNA syndrome", "Short-lasting unilateral neuralgiform headache attacks with cranial autonomic symptoms",
"Paroxysmal hemicrania",
"Hypnic headache",
"Vestibular migraine",
"Abdominal migraine",
"Status migrainosus",
"New daily persistent headache", "NDPH",
"Psychomotor retardation",
"Catatonia",
"Tropical spastic paraparesis", "HTLV-1 associated myelopathy",
"Diabetic amyotrophy",
"Diabetic autonomic neuropathy",
"Diabetic radiculoplexus neuropathy",
"Acute hemorrhagic leukoencephalitis", "Hurst disease",
"Neurologic paraneoplastic syndromes",
"Omenn syndrome (neurological complications)",
"Acute flaccid paralysis",
"Spinal cord injury", "SCI",
"Anoxic brain injury", "global hypoxic-ischemic injury",
"Cerebral edema",
"Brain herniation syndromes",
"Sundowning (Alzheimer's-related)",
"Pellagra (niacin deficiency encephalopathy)",
"Leprosy (Hansen's disease, neuropathy)",
"Tropical ataxic neuropathy",
"Konzo (epidemic spastic paraparesis)",
"Lathyrism",
"Pesticide-induced neuropathy",
"Organophosphate neuropathy",
"Carcinomatous meningitis", "leptomeningeal carcinomatosis",
"Gliomatosis cerebri",

"POEMS syndrome", "Polyneuropathy, Organomegaly, Endocrinopathy, M-protein, Skin changes",
"Waldenström macroglobulinemia (neurological)",
"Monoclonal gammopathy of undetermined significance neuropathy", "MGUS neuropathy",
"CRION (chronic relapsing inflammatory optic neuropathy)",
"Neuromyelitis optica spectrum disorder", "NMOSD",
"MOG antibody disease", "myelin oligodendrocyte glycoprotein disorder",
"Autoimmune encephalitis", "anti-NMDA receptor encephalitis",
"Limbic encephalitis",
"Basal ganglia encephalitis",
"Voltage-gated potassium channel complex antibody encephalitis", "VGKC encephalitis",
"Glutamic acid decarboxylase antibody neurologic syndromes", "GAD antibody syndromes",
"CASPR2 antibody encephalitis", "Contactin-associated protein-like 2 encephalitis",
"Morvan syndrome",
"PANDAS (pediatric autoimmune neuropsychiatric disorders associated with strep)",
"Sydenham chorea", "rheumatic chorea",
"Chorea gravidarum",
"Hemiballismus",
"Neuroferritinopathy",
"Benign hereditary chorea",
"Dentatorubral-pallidoluysian atrophy", "DRPLA",
"Spastic dystonia",
"Kayser-Fleischer ring (Wilson's disease sign)",
"Oromandibular dystonia",
"Blepharospasm",
"Cervical dystonia", "torticollis",
"Writer's cramp", "focal hand dystonia",
"Alien hand syndrome",
"Conduction aphasia",
"Broca's aphasia", "expressive aphasia",
"Wernicke's aphasia", "receptive aphasia",
"Global aphasia",
"Prosopagnosia", "face blindness",
"Balint's syndrome",
"Gerstmann syndrome",
"Phineas Gage syndrome", "frontal lobe syndrome",
"Temporal lobe syndrome", "Kluver-Bucy syndrome",
"Papez circuit lesions", "amnestic syndromes",
"Transient epileptic amnesia",

"Psychogenic amnesia",
"Electroconvulsive therapy-induced amnesia",
"Mild cognitive impairment", "MCI",
"Disorders of consciousness", "coma", "vegetative state", "minimally conscious state",
"Brain death",
"Hypersomnia",
"Kleine-Levin syndrome", "recurrent hypersomnia",
"Periodic limb movement disorder", "PLMD",
"Sleep apnea", "obstructive sleep apnea", "OSA", "central sleep apnea", "CSA",
"Cataplexy",
"Circadian rhythm sleep disorder",
"Somnambulism", "sleepwalking",
"Night terrors", "pavor nocturnus",
"Insomnia",
"Parasomnias",
"X-linked adrenoleukodystrophy", "X-ALD",
"Spinal cord infarction",
"Spinal arteriovenous malformation",
"Scoliosis (neurological involvement)",
"Hydromyelia",
"Diastematomyelia",
"Basilar invagination",
"Foramen magnum stenosis",
"Kugelberg-Welander disease (SMA type III)",
"Aran-Duchenne muscular atrophy",
"Scapuloperoneal spinal muscular atrophy",
"Emery-Dreifuss muscular dystrophy",
"Myotonia congenita", "Thomsen disease", "Becker type",
"Facioscapulohumeral muscular dystrophy", "FSHD",
"Limb-girdle muscular dystrophy", "LGMD",
"Distal muscular dystrophy",
"Oculopharyngeal muscular dystrophy", "OPMD",
"Inclusion body myositis", "IBM",
"Necrotizing myopathy",
"Infectious myositis",
"Acute rhabdomyolysis with neuropathy",
"Episodic ataxia",
"Spasmodic dysphonia", "laryngeal dystonia",
"Tardive dystonia",
"Septo-optic dysplasia",
"Joubert syndrome",
"Dandy-Walker syndrome",

"Arachnoid cyst",
"Colloid cyst",
"Craniopharyngioma",
"Pineal tumor", "pinealoma",
"Pituitary adenoma (neurological involvement)",
"Hypothalamic hamartoma",
"Choroid plexus papilloma",
"Hydranencephaly",
"Holoprosencephaly",
"Schizencephaly",
"Lissencephaly",
"Pachygyria",
"Polymicrogyria",
"Heterotopia (neuronal migration disorder)",
"Dural arteriovenous fistula",
"Cerebral venous sinus thrombosis", "CVST",
"Budd-Chiari syndrome (neurological involvement)",
"Central pontine myelinolysis", "osmotic demyelination syndrome",
"Extrapontine myelinolysis",
"Radiation myelopathy",
"Iatrogenic neurological disorders",
"Hemicraniectomy complications",
"Hyperperfusion syndrome (post-carotid endarterectomy)",
"Scorpion venom neuropathy",
"Snake bite neurotoxicity",
"Tick paralysis",
"Spider envenomation (e.g., black widow neurotoxicity)",
"Shellfish toxin-induced neuropathy", "paralytic shellfish poisoning",
"Ciguatera poisoning (neurological)",
"Scombroid poisoning",
"Thrombotic thrombocytopenic purpura (neurological involvement)", "TTP",
"Shaken baby syndrome", "abusive head trauma"